\documentclass[lettersize,journal]{IEEEtran}
\usepackage{amsmath,amsfonts}
\usepackage{amsmath}
\usepackage{amssymb}
\usepackage{tabularx} 
\usepackage{algorithm}
\usepackage{algorithmicx}
\usepackage{array}
\usepackage[caption=false,font=normalsize,labelfont=sf,textfont=sf]{subfig}
\usepackage{textcomp}
\usepackage{stfloats}
\usepackage{url}
\usepackage{verbatim}
\usepackage{graphicx}
\usepackage{cite}
\usepackage{algpseudocode}
\usepackage{booktabs}
\usepackage{balance}
\usepackage{glossaries}
\makeglossaries
\usepackage{multirow}
\usepackage{xcolor}
\usepackage{hyperref}
\hypersetup{
	colorlinks=true,
	linkcolor=blue,       
	citecolor=blue,    
	filecolor=blue,      
	urlcolor=blue         
}

\begin{document}

\title{Extreme Precipitation Nowcasting using Multi-Task Latent Diffusion Models}

\author{Chaorong Li,~\IEEEmembership{Member,~IEEE, Xudong Ling, Qiang Yang, Mingxiang Chen, Fengqing Qin,Yuanyuan Huang}
        
\thanks{This work is supported by Science and Technology Department of Sichuan 	Province(No.2024ZYD0089);Yibin University Science and Technology(No.2024XJYY03). \emph{(Corresponding author: C.R.~Li, e-mail: lichaorong88@163.com; C.R. Li and X.D. Ling contributed equally to this work.)}}
	
\thanks{C.R. Li , X.D.Ling, Q.Yang, M.X. Chen and F.Q. Qin are with the School of Computer Science and Technology, Yibin University, Yibin 644000, China.} 
\thanks{Y.Y. Huang is with the School of Artificial Intelligence, Chengdu University of Information Technology, chengdu  610225, China.}
 }



\maketitle

\begin{abstract}
	\textcolor{black}{Deep learning models have achieved remarkable progress in precipitation prediction. However, they still face significant challenges in accurately capturing spatial details of radar images, particularly in regions of high precipitation intensity. This limitation results in reduced spatial localization accuracy when predicting radar echo images across varying precipitation intensities. To address this challenge, we propose an innovative precipitation prediction approach termed the Multi-Task Latent Diffusion Model (MTLDM). The core idea of MTLDM lies in the recognition that precipitation radar images represent a combination of multiple components, each corresponding to different precipitation intensities. Thus, we adopt a divide-and-conquer strategy, decomposing radar images into several sub-images based on their precipitation intensities and individually modeling these components. During the prediction stage, MTLDM integrates these sub-image representations by utilizing a trained latent-space rainfall diffusion model, followed by decoding through a multi-task decoder to produce the final precipitation prediction. Experimental evaluations conducted on the MRMS dataset demonstrate that the proposed MTLDM method surpasses state-of-the-art techniques, achieving a Critical Success Index (CSI) improvement of 13-26\%. }
\end{abstract}

\begin{IEEEkeywords}
Precipitation nowcasting, Extreme Precipitation, Diffusion model, Multi-task.
\end{IEEEkeywords}

\section{Introduction}
Short-term precipitation is difficult to forecast accurately due to its rapid formation and dissipation. Short-term forecasts, i.e., 1–2 h in advance, can capture the position of weather systems and predict the range of precipitation in a timely manner, which is crucial for promptly informing the public about the time, location, and intensity of rainfall\cite{ebert2003wgne,campbell2005weather}. However, as global climate change and urbanization accelerate, precipitation patterns are constantly changing, which poses higher demands on precipitation prediction technology. Deep learning models\cite{ravuri2021skilful,agrawal2019machine,akbari2018short} have been shown to have a more powerful predictive performance than numerical weather prediction (NWP)\cite{al2010review}. However, as mentioned above, deep learning models still have serious deficiencies in the accuracy of spatiotemporal information prediction, making it difficult for them to meet practical needs. Therefore, researching high-precision forecasting models not only has important scientific value but also has broad application prospects.
The inaccuracy of spatiotemporal information prediction is mainly manifested in the blurring of radar echo images generated during the prediction process, which is the result of inadequate convergence during the training process due to imperfect design structures of deep learning models. Moreover, the poor predictive performance of the details of spatial features is reflected in the low critical success index (CSI) of the generated radar echo images. Currently, deep learning models can hardly exceed a CSI of 0.5 under heavy precipitation conditions ($\geq$16 mm/h)\cite{ravuri2021skilful,zhang2023skilful}. This situation indicates that deep learning models have significant problems in predicting the detailed features of images and cannot accurately predict precipitation conditions. As the prediction duration increases, the spatial prediction accuracy decreases dramatically. The inaccuracy of long sequence prediction is one of the main problems faced in all current deep learning research. Existing studies and our experiments clearly show that when the prediction duration exceeds 1 h, the CSI indicators for both heavy and light precipitation decrease sharply.

Currently, the mainstream generative backbone network is the encoder–decoder (E–D) structure,  such as  Trebing et al.\cite{trebing2021smaat} introduced the Self-Attention(SA) into UNet and reduced the number of parameters in convolution-based models. In 2021, Ravuri Name et al. \cite{ravuri2021skilful}  achieved breakthrough performance using generative adversarial networks (GANs) to train a Convolutional Gated Recurrent unit (ConvGRU)-based E–D model (called DGMR) for precipitation prediction. This research attracted widespread attention to the use of deep networks in the field of precipitation prediction (i.e., weather prediction). Ling et al. \cite{ling2024tu2net} proposed a nested temporal UNet (TU2Net) that deepens the network hierarchy, significantly improving the quality of the predicted spatiotemporal information compared to the DGMR. Since deep networks are data-driven learning methods and cannot adhere to inherent physical laws, Zhang et al. \cite{zhang2023skilful} proposed NowcastNet, a nonlinear nowcasting model for extreme precipitation, which unifies physical evolution schemes and conditional learning methods into an end-to-end E–D network. However, like DGMR, NowcastNet still has low CSI values for heavy rainfall, and the CSI value for heavy precipitation($\geq$64 mm/h) decreases sharply to around 0.1 in the first 30 min.

The diffusion probabilistic model (DPM) is a generative model that has emerged in the past two years and has been widely applied in video generation, image segmentation, knowledge reasoning, and other applications \cite{ho2020denoising, feng2024latent}. It has been proven to outperform GAN models in various application areas. However, literature on applying the DPM to precipitation prediction only began to appear at the end of 2023. Zhihan et al. \cite{gao2024prediff} proposed using the DPM for precipitation prediction, and its CSI value outperforms that of the GAN-based DGMR (both methods use UNet as the backbone network). Zhao et al.\cite{zhao2024advancing} combined the transformer with the DPM for precipitation prediction, confirming that the transformer achieves better CSI values than UNet. However, as the DPM has just begun to be applied to weather prediction such as rainfall, there is still significant room for improvement in its predictive performance. For instance, experimental data \cite{zhao2024advancing} show that when the precipitation is $>$ 3.5 mm/h, the CSI values of all of the methods become very low in heavy precipitation areas, and their average values rarely exceed 0.3. \textcolor{black}{The fundamental reasons why existing deep learning models perform poorly under heavy precipitation conditions can be attributed to two aspects: data characteristics and model structure.
From a data perspective, heavy precipitation events occur less frequently in nature, resulting in a serious shortage of heavy precipitation samples in the training data\cite{ravuri2021skilful,zhang2023skilful}, resulting in an imbalanced data distribution problem. This imbalance causes the model to tend to optimize the prediction of common precipitation intensities during training, at the expense of accurately capturing rare heavy precipitation events. In addition, heavy precipitation usually exhibits high nonlinearity and uncertainty in time and space, making it particularly difficult to learn these complex patterns from limited samples.
From the perspective of model structure, existing deep learning models with single-task architectures have difficulty in simultaneously processing the feature representation of precipitation of different intensities\cite{sonderby2020metnet}. Light precipitation and heavy precipitation show significantly different patterns and evolution laws on radar echo maps, and a single network architecture has difficulty capturing these differentiated features at the same time. In addition, traditional models usually use global loss functions such as mean square error (MSE)\cite{trebing2021smaat,ling2024spacetime}. When faced with images with a small proportion of heavy precipitation, such functions tend to smooth the prediction results and cannot accurately reconstruct the spatial details and boundary information of high-intensity precipitation areas. This results in the CSI value in heavy precipitation areas being significantly lower than that in light precipitation areas, which seriously limits the practical value of the model in extreme weather warning.}

\begin{figure}[htbp]
	\centering
	\includegraphics[width=0.48\textwidth]{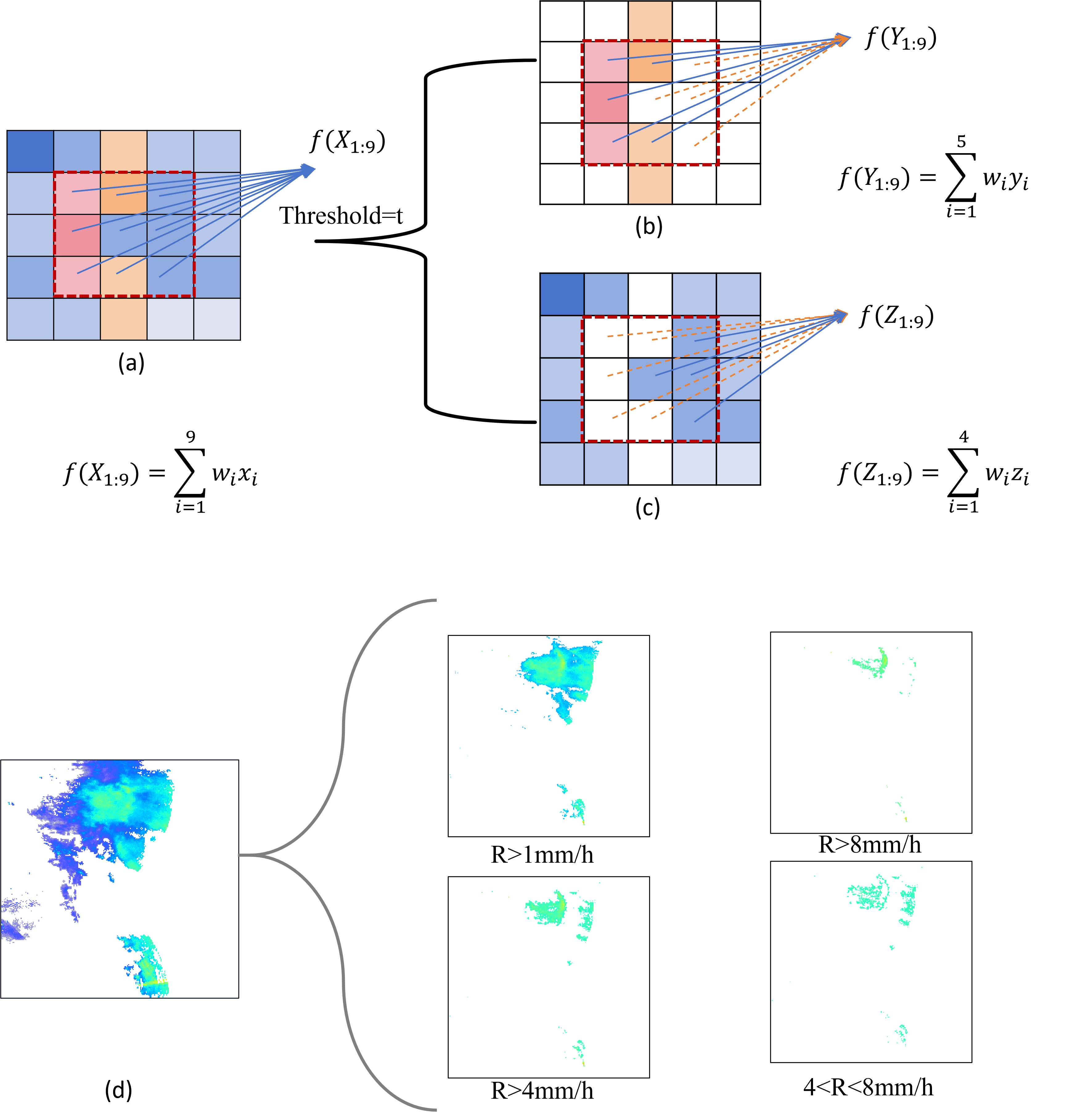}
	\caption{Radar threshold decomposition and feature extraction. (a) Original radar image, and (b and c) radar sub-images. \textcolor{black}{(d) Decomposition of real radar image according to precipitation threshold (R represents precipitation)}}
	\label{fig1}
\end{figure}

To improve the spatiotemporal aceuracy of precipitation prediction, we adopteda hierarchical prediction method that decomposes complex problems, thus breaking down the precipitation image into different components.  We use threshold decomposition technology to divide radar images into multiple sub-images representing different precipitation amounts, each of which corresponds to a specific component in the original image. When using deep neural networks to extract features, the network processes local or overall pixel features of radar images. By extracting features from each sub-image separately, the interference from other precipitation level images can be effectively shielded, thereby improving the accuracy of feature extraction and simplifying the model training process. This method helps reduce noise interference and allows the model to focus more on task learning at different precipitation intensity levels.  Fig.\ref{fig1} presents a schematic diagram of threshold decomposition and feature extraction of radar images.
 
\begin{figure}[htbp]
	\centering
	\includegraphics[width=0.48\textwidth]{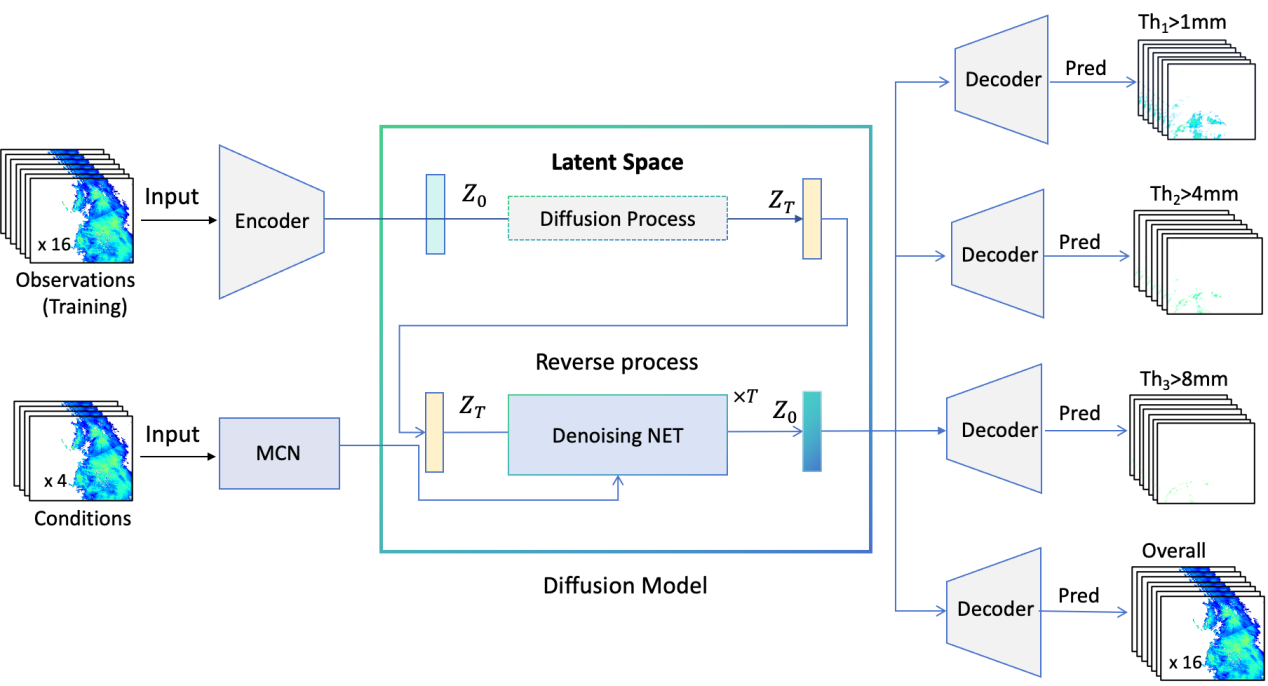}
	\caption{Overview of MTLDM for Precipitation Prediction.}
	\label{fig2}
\end{figure}

 Fig.\ref{fig2}  shows overview of the proposed multi-task latent diffusion model (MTLDM) for precipitation prediction. After the input radar image sequence is predicted by the latent space diffusion model, different decoders are used to generate different precipitation sub-images from the predicted latent variables. Simultaneously, we also predict and generate the overall image. Based on the radar precipitation intensity, we set three thresholds ($Th_1$, $Th_2$, and $Th_3$) to segment a single radar image frame into different precipitation sub-images to train the Denoising Net and use a multi-task structure to predict the radar images of different precipitation intensities.

This approach helps to better understand and predict the detailed situations of rainfall. This decomposition can be achieved through various methods, such as texture-structure-based decomposition \cite{zhao2023learning}, low-rank-sparse-based decomposition \cite{peng2021low}, and statistical method-based decomposition \cite{rohman2021through}. We investigated the use of a decomposition method based on precipitation intensity thresholds to classify processed precipitation data according to the intensity to obtain precipitation amounts at different intensity levels and analyze the spatial distribution and movement patterns of precipitation at each intensity level. By decomposing the precipitation intensity and precipitation motion characteristic distributions, we treated the decomposed images as subtasks for prediction. This allowed us to more precisely target specific tasks to improve the accuracy of the different precipitation predictions, thereby also improving the CSI.

Traditional deep learning models\cite{shi2017deep,ma2023mm,ma2024db,luo2022predrann,luo2022experimental} have struggled to maintain spatial accuracy in radar echo images, particularly in regions of intense precipitation, leading to misalignment and diffusion in their forecasts. MTLDM overcomes these limitations by decomposing radar images into distinct components based on varying precipitation intensities and using task-specific networks to predict each component independently. This divide-and-conquer approach enhances both spatial and temporal consistency in precipitation predictions.

\textcolor{black}{In summary, the main contributions of the Multi-Task Latent Diffusion Model (MTLDM) proposed in this paper are as follows:}

\begin{itemize}
\item  \textcolor{black}{We proposed a "divide and conquer" strategy based on decomposition technology, decomposing radar images into sub-images corresponding to different precipitation intensities according to rainfall intensity, and designed a dedicated decoding task for each sub-image, so that the decoder can more accurately capture the spatial characteristics of precipitation areas with different intensities.}
\item \textcolor{black}{Based on the latent diffusion framework, MTLDM directly performs the diffusion process in the compressed latent space and directly decodes the generated latent space results. For the trained prediction model, there is no need to additionally train the latent space diffusion prediction model, which significantly reduces the research cost.}
\item \textcolor{black}{The multi-task learning framework we designed can optimize multiple related objectives simultaneously, significantly improving the prediction performance in high-intensity precipitation areas ($\geqslant 8 mm/h$) while maintaining the prediction accuracy in low-intensity precipitation areas.}

\end{itemize}

\section{RELATED WORKS}
In this section, we briefly introduce the research on precipitation prediction based on diffusion model.
 \subsection{Denoising diffusion probabilistic model (DDPM)}
Denoising diffusion probabilistic model (DDPM) \cite{ho2020denoising} is a generative model used for creating image sequences by reversing a gradual noise addition process, based on probabilistic principles. The use of DDPM includes two main stages: training and sampling.

\subsubsection{\textbf{Training Process} }

The training process of a diffusion model typically involves learning the reverse process, which progressively restores a noisy image back to a clean one. The steps are as follows:

\textbf{Random Sampling of Initial Image}. 
Sample a clean image \( x_0 \) from the real image data distribution.

\textbf{Forward Diffusion Process}. 
A forward diffusion process gradually transforms the clean image \( x_0 \) into a sequence of progressively noisier images \( x_1, x_2, \dots, x_T \) by adding Gaussian noise. This process can be represented as a parametric Gaussian process where noise \( \epsilon \) is drawn from the standard normal distribution \( \mathcal{N}(0, I) \). The process is defined as:
$x_t = \sqrt{\alpha_t} x_0 + \sqrt{1 - \alpha_t} \epsilon
$, where \( \alpha_t \) is a hyperparameter controlling the noise level.

\textbf{Learning the Reverse Diffusion Process}. 
Train a model \( \epsilon_\theta(x_t, t) \) to predict the noise \( \epsilon \) at each timestep \( t \), in order to denoise the noisy image \( x_t \) back to \( x_0 \). The objective is to reverse the diffusion process by predicting the noise and progressively denoising the image. The loss function at each timestep is typically the mean squared error (MSE) of the predicted noise:
$L(\theta) = \mathbb{E}_{x_0, \epsilon, t} \left[ \lVert \epsilon - \epsilon_\theta(x_t, t) \rVert^2 \right]
$, where \( \epsilon \) is the true noise sampled from the standard normal distribution, and \( \epsilon_\theta(x_t, t) \) is the model's predicted noise.

\textbf{Multi-Step Reverse Process}. 
During the training process, the model gradually denoises the image from timestep \( T \) back to timestep \( 0 \). The reverse diffusion process at each timestep follows the equation: $	x_{t-1} = \frac{1}{\sqrt{\alpha_t}} \left( x_t - \frac{1 - \alpha_t}{\sqrt{1 - \alpha_t}} \epsilon_\theta(x_t, t) \right)
$. This equation updates the image at each step based on the model's noise prediction until a clean image \( x_0 \) is restored.

\textbf{Training Objective}. 
The objective is to optimize the noise prediction function \( \epsilon_\theta \) by minimizing the noise prediction error, ensuring that the reverse diffusion process progressively removes noise and restores the clean image.

\subsubsection{\textbf{The sampling prediction process}}
The sampling prediction process of the diffusion model generates an initial noisy image $x_T$ from the standard normal distribution $N(0, I)$ and then gradually denoises it through the inverse diffusion process. At each time step, the model predicts the noise component $\epsilon_\theta(x_t, t)$ based on the current image $x_t$ and time $t$. The denoising step adjusts the image to remove noise using the equation $ x_{t-1} = \frac{1}{\sqrt{a_t}} \left( x_t - \frac{1 - \bar{a}_t}{\sqrt{1 - a_t}} \epsilon_\theta (x_t, t) \right)$ and adds a small amount of noise except in the last step. This process continues until the final realistic image $x_0$ is generated.

 \subsection{Precipitation prediction based on DDPM}
In the task of precipitation forecasting, diffusion models have been widely applied to improve prediction accuracy and address the limitations of traditional deep learning methods. In recent years, several research efforts have focused on integrating diffusion models with different spatiotemporal information processing mechanisms to enhance the timeliness and accuracy of precipitation nowcasting.

Gao et al.\cite{gao2024prediff} proposed a conditional latent diffusion model named PreDiff for probabilistic spatiotemporal forecasting. This approach introduces a knowledge alignment mechanism constrained by physical laws, evaluating deviations from these constraints at each denoising step and adjusting the transition distribution accordingly. This ensures that the predictions not only capture uncertainty but also remain consistent with physical laws. This method significantly improves upon the generation of blurred and physically implausible predictions, demonstrating strong practicality and accuracy in handling complex weather systems. To further enhance sampling efficiency and prediction quality, Zhao et al.\cite{zhao2024advancing}  developed a diffusion model based on a spatiotemporal transformer. This model employs a rapid diffusion strategy to effectively address the long sampling times typical of diffusion models and incorporates a spatiotemporal transformer-based denoiser to replace traditional U-Net denoisers, significantly improving the quality of precipitation nowcasting. Compared to traditional methods, this model not only enhances meteorological evaluation metrics but also delivers clearer and more accurate predictions in typical weather events such as thunderstorms.

In addressing issues such as blurriness and mode collapse in existing generative models, Nai et al.\cite{nai2024reliable} introduced a probabilistic diffusion-based precipitation forecasting method. This method learns a series of neural networks to reverse the predefined diffusion process, generating the probability distribution of future precipitation fields. Experiments show that this method performs well in uncertainty evaluation metrics such as Continuous Ranked Probability Score (CRPS) and spread-skill ratio, further demonstrating the potential of diffusion models to improve the reliability of uncertainty prediction.

Furthermore, Yu et al.\cite{yu2024diffcast} proposed a unified framework based on residual diffusion (DiffCast), which models the global deterministic motion and local stochastic variations of precipitation systems separately. This effectively tackles the chaotic nature of complex weather systems. The model is flexible enough to be adapted to any spatiotemporal model and has demonstrated superior prediction accuracy in experiments on multiple public radar datasets. To improve long-term precipitation forecasting accuracy, Ling et al.\cite{ling2024two} designed a two-stage rainfall forecasting diffusion model (TRDM). This model captures temporal information under low-resolution conditions in the first stage, and reconstructs high-resolution spatial information in the second stage, effectively balancing the performance between temporal and spatial modeling and significantly improving the accuracy of long-term precipitation forecasts.

Recently, Ling et al.\cite{ling2024spacetime} further proposed a spacetime separable latent diffusion model with intensity structure information (SSLDM-ISI, for simplicity, it is referred to as SSLDM). This model efficiently extracts and integrates spatiotemporal information through a Spatiotemporal Conversion Block (STC Block) and introduces a latent space encoding technique based on rainfall intensity structural information, enhancing the representation capability of extreme rainfall events. Experiments on various datasets show that SSLDM outperforms existing advanced technologies in both meteorological evaluation and image quality evaluation metrics, particularly excelling in short-term precipitation forecasting tasks. This work further enriches the application of diffusion models in spatiotemporal information processing, especially in predicting extreme rainfall events, showcasing high application value.
  
\section{Method}
\subsection{Diffusion Probabilistic Models based on Radar Images}
 Our model uses the radar image \(Y = \{y_{1}, \cdots, y_{M}\}\) of the first \(M\) frames at time \(S\) to predict the radar image \(X = \{x_{1}, \cdots, x_{N}\}\) of \(N\). The input radar image sequence is segmented using thresholds and is decomposed into multiple components, each of which is fed into a different diffusion model. Each diffusion model maps the input image to the latent space through an encoder, performs a forward diffusion process, uses a spatiotemporal separable network to generate multiple intermediate latent representations, and finally obtains the final state representation. For each independent diffusion model, the prediction formula is as follows:

\begin{equation}
	\begin{split}
		P(X^i\mid Y^i) &\sim P(z_0\mid Y^i)\\
		&=\int P(z_{0:T}\mid Y^i,\theta_E,\theta_D,\theta_C,\theta_\epsilon)\,dz_{1:T}
	\end{split}
\end{equation}
 where \(\theta_{E}, \theta_{D}, \theta_{C}, \text{ and }\theta_{\epsilon}\) are the parameters of the encoder, decoder, conditional encoder, and denoising backbone network, respectively. \(z_{1:T}\) denotes the latent vectors of the same dimensionality as the data \(z_{0}\). \(z_{0} = E(X^{i})\) encodes the image into the latent space. Then, new images can be generated by sampling representations \(z\) from the diffusion model and subsequently decoding them into images using the learned decoder \(X^{i} = D(z_{0})\). Four consecutive radar observations are used as the conditions for the dispersion model, i.e., \(M = 4\), which allows sampling of multiple realizations of future precipitation. Each realization is \(16\) frames, i.e., \(N = 16\).

 The four frames of the radar image input and output are first compressed into the latent space by the encoder. The latent space contains almost all of the original radar image information, so this information will not be lost after the decoder. The real prediction is carried out by the diffusion model in the latent space. The predicted coded information is decoded by different threshold decoders, and the corresponding precipitation distribution maps with different thresholds are output. The proposed multi-task hidden space diffusion model has multiple prediction outputs. To clearly express the benefits of the image decomposition, the prediction method for all of the radar maps is referred to as the \textit{MTLDM (oa)}, and the prediction method for different radar maps under different thresholds is referred to as the \textit{MTLDM}. 

In MTLDM, we use the latent diffusion model (LDM) as the generative framework because the LDM has several advantages over diffusion models (DMs): the LDM can generate images with a higher quality and more detail than DMs. This is because the LDM works in latent space, which allows it to capture more complex and nuanced features of the data. Second, the LDM is more computationally efficient than DMs. This is because the LDM can generate images from lower-resolution noise, which requires fewer computations.

Each diffusion model maps 16 frames of observations to the latent space through an encoder, undergoes a forward diffusion process, and uses a spatiotemporal separable network to generate multiple intermediate latent representations, ultimately obtaining the representation of the final state. These latent representations are processed in combination with specific conditional information. We split the four frames of the context information into components (referred to as the conditions) and feed them into the encoder to obtain the latent space conditional information, which is then input into the Spacetime Separable NET to control the predicted output. After completing the prediction in the latent space, the processed latent representations are restored back to the pixel space through a decoder, generating the predicted 16 frames of the radar images. In this way, the model can efficiently capture the intensity structure information of the input images, improving the accuracy of the precipitation prediction. Its overall architectural design enables the model to efficiently capture the intensity structure information of the input images, thereby improving the accuracy of the precipitation prediction.

\subsection{Model details}
\begin{figure*}[htbp]
	\centering
	\includegraphics[width=0.9\textwidth]{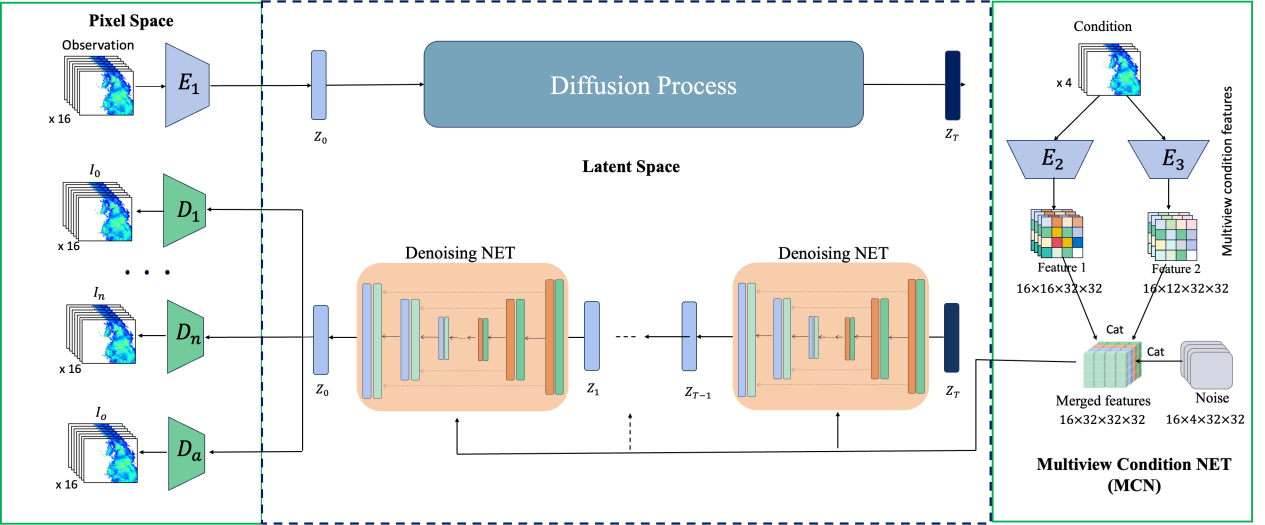}
	\caption{The detailed architecture of the Multi-task Latent Diffusion Model (MTLDM).}
	\label{fig3}
\end{figure*}
 Fig. \ref{fig3} shows the details of the MTLDM precipitation forecast model, which shares the same backbone network with the spacetime separable latent diffusion model (SSLDM, which is our previous work)\cite{ling2024spacetime}. The condition consists of four frames of radar images, which are used to predict 16 frames of radar images. The diffusion model adopts the latent diffusion model \cite{rombach2022high} meaning that the diffusion prediction encodes the input components (referred to as the conditions) and performs prediction in the latent space and then finally restores the predicted radar images through a decoder. The conditions are encoded through multiple encoders called Multiview Condition Net, forming multi-perspective features, which are then concatenated with Gaussian noise and fed into the Denoising NET. In the latent space encoding, the intensity structure information about the radar images is considered and is used to highlight the high-frequency information of the image and to enhance the prediction accuracy for areas of heavy rainfall.
 
\textbf{Input Processing}: The left side of the image shows the input sequence of the radar images, labeled as Observation. Each input sequence contains 16 frames of radar images.

\textbf{Pixel Space to Latent Space}: In each diffusion model, the images are first mapped from the pixel space to the latent space using the intensity structure information. This mapping process is completed through encoder , which converts the input images into an initial latent representation $z_0$.

\textbf{Forward Diffusion Process}: The initial latent representation $z_0$ enters the forward diffusion process and is gradually transformed into the final latent representation $z_T$. 

\begin{figure}[htbp]
	\centering
	\includegraphics[width=0.5\textwidth]{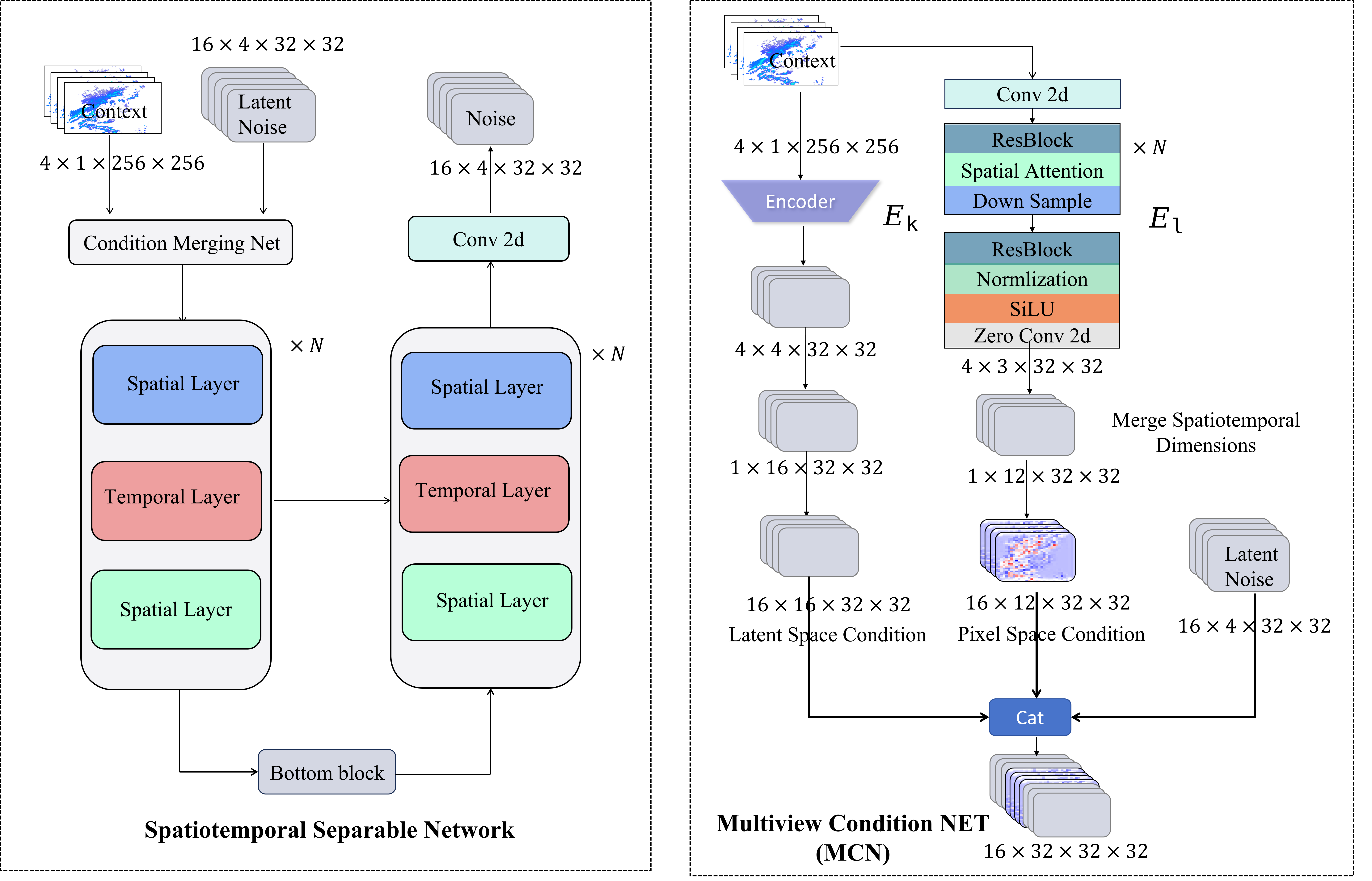}
	\caption{Denoising NET and multiview condition NET. (a)Denoising NET,(b)Multiview condition NET}
	\label{fig4}
\end{figure}

\textbf{Denoising Network}: During the diffusion process, the latent representation $z_0$ is processed using several steps in spatiotemporal separable denoising networks (Fig. \ref{fig4}(a)), and each step generates an intermediate latent representation $z_i , i=1,2,\cdots, T-1$. Finally, the final state representation $z_T$ is generated.

\textbf{Condition Handling}: The condition module is generated by the encoder of the input radar image and the context information. The context information is obtained by decomposing the input reduced-dimensional image sequence (each sequence contains four images). The context information and the conditions are combined to form the latent space condition.

\textbf{Decoding and Prediction}: The processed latent representation is decoded back to the pixel space by the decoder ($D_1$-$D_n$, and $D_a$) to generate a predicted radar image sequence. Among them, $D_1$-$D_n$ decode the predicted latent space information into radar images at different thresholds, while $D_a$ decodes the overall radar image directly, without threshold segmentation. This process is completed by multiple decoders, which are used to decode and restore the radar images of $n$ different precipitation intensities.  It is worth noting that the output of $D_a$ is the output of the model MTLDM(oa), and its performance is comparable to that of SSLDM\cite{ling2024spacetime}. The decoder is the decoding part of the AutoEncoder. Although the inputs of these decoders are different precipitation maps, they have the same structure.

\textbf{Multi-Task Learning}: \textcolor{black}{In order to ensure that the diffusion prediction model predicts in the same latent space distribution and verify the performance advantage of the multi-task decoder, we fixed the parameters of the encoder and only updated the parameters of the decoder. The input of the multi-task decoder is the decomposed threshold image, and its training goal is to accurately reconstruct the image. Since the encoder parameters remain unchanged and the input condition of the MCN is still the undecomposed radar image, the trained latent space prediction model can run normally. For the predicted latent space information, we directly use the multi-task decoder to decode it to obtain the final result.}

\textbf{AutoEncoder}:We use the Autoencoder used by Rombach et al.\cite{rombach2022high} and the intensity structural information of the radar to encode the latent space variables \cite{ling2024spacetime}. This module uses a ResnetBlock stack based on a convolutional network, and the KL-enalty factor is used during the training. To improve the reconstructing quality of the encoder, we added the structural similarity index similarity (SSIM)\cite{wang2004image} and learned perceptual image patch similarity (LPIPS) \cite{zhang2018unreasonable} losses to train the network. Unlike traditional methods, the SSIM evaluates the image brightness and contrast, as well as the structural information. Using the SSIM and LPIPS as loss functions, high-frequency signals in extreme precipitation events can be captured and reconstructed better, maximizing the performance of the latent of the DM. Thus, the AutoEncoder loss can be expressed as follows:
\begin{equation}
	L = L_{SSIM} + L_{KL} + L_{LPIPS} + L_{\|x - x'\|}	
\end{equation}
For the detailed implementation of the autoencoder, please refer to our previous work\cite{ling2024spacetime}. We use AutoEncoder for the $E_1$ and $E_2$ of MTLDM, which encode the radars into latent space.

\textbf{Multiview Condition NET}: Fig. \ref{fig4}(b) shows the detailed framework of the multiview condition NET. It takes a conditional radar frame with size $4\times1\times256\times256$ as the input. The multiview condition NET consists of two branches . The left branch obtains the latent space condition through the latent space encoder($E_2$ ), and its structural parameters remain unchanged during the diffusion model training process. The right branch uses the pixel encoder to encode the original image space($E_3$). The main part of the network uses the spatial attention mechanism and the down-sampling layer to reduce the spatial scale of the condition to $1\times12\times32\times32$ to obtain the pixel space condition.

The original representation dimension of the latent space condition is $(T, C, W, H)$, where $T$ is the number of time steps, $C$ is the number of channels, and $W$ and $H$ are the spatial scales. To better model the time dimension, we merge the channel dimension $C$ and the time dimension $T$ into a new dimension $C'$ and replicate it 16 times in the time dimension to obtain a spatiotemporal fusion latent space vector of size $(T=16, C=16, W=32, H=32)$. Similarly, we replicate the pixel space condition 16 times in the time dimension to obtain a spatiotemporal condition representation with dimensions $(T=16, C=12, W=32, H=32)$.

\section{Experimental results and Analysis}
\subsection{Datasets and evaluation metrics}
\textcolor{black}{The experiments were conducted on two datasets. The multi-radar multi-sensor (MRMS) dataset for the United States\cite{ravuri2021skilful} provided data from 2020–2021 for training and data from 2022 for testing. The Swedish dataset (SW)\cite{ling2024spacetime} includes radar data from February 6, 2017, to November 12, 2021, with data from February 2017 to November 2020 used for training, and the remaining data used for validation and testing. The SW dataset is formatted similarly to grayscale images, with values ranging from 0 to 255, where 0 represents no echo, and 255 represents no data (indicating either a measurement point within radar coverage but with no reported echo, or areas outside the radar coverage). For both datasets, we cropped the images from the radar stream to a size of 256×256 pixels.}
\subsection{Baselines}
We compared several models, including the Python framework for short-term ensemble prediction system (PySTEPS) (a popular machine learning method for precipitation prediction) \cite{pulkkinen2019pysteps}, the DGMR model (a well-known conditional GAN for short-term precipitation prediction) and the timed racing diffusion model (TRDM) (a two-stage diffusion model-based method for precipitation prediction) \cite{ling2024two}. The TRDM includes two versions: one that uses a super-resolution diffusion model for prediction in the second stage (called TRDMv1), and another that uses LDM for prediction (called TRDMv2).The parameter sizes of these models are presented in Table \ref{tab:model_size}. 
Two important metrics for evaluating precipitation are used: the CSI \cite{schaefer1990critical, ravuri2021skilful,zhang2023skilful} and the continuous ranked probability score (CRPS)\cite{ravuri2021skilful,gneiting2007strictly}. The CSI is a commonly used statistical indicator for evaluating the accuracy of categorical forecasts and is often used to assess the prediction accuracy of weather events such as precipitation, storms, and snowfall. In evaluation, a higher CSI value indicates a higher accuracy of categorical forecasts. In precipitation assessment, the CRPS helps evaluate the consistency between the predicted precipitation probability distribution and the actual observed rainfall. Its advantage is that it considers both the precision and uncertainty of the predictions, making it particularly suitable for evaluating the performances of meteorological models. A smaller CRPS value indicates a better consistency between the predicted probability distribution and the observed values, suggesting a higher prediction accuracy.

\subsection{Diffusion Model Training and Sampling Overview}
The diffusion model training involves several steps. Firstly, initial data is sampled from the real data distribution. Then, a time step in the diffusion process is randomly selected, and noise $\epsilon$ is sampled from the standard normal distribution. The difference between the model prediction noise and the actual noise is calculated and minimized. Model parameters are updated via gradient descent, and this process is repeated until the model converges, enhancing the model's ability to recover the original data from noise. The training is carried out using the PyTorch 1.3.1 framework on 10 A6000 GPU devices under Ubuntu 22.04(\textcolor{black}{Detailed hyperparameter settings are shown in Table \ref{trainingtab}}). In the inference phase, the DDPM sampling method is used on a single A800 GPU. It takes approximately 480 seconds to sample a batch of data, which means generating 16 frames of prediction results. For the diffusion model and denoising network, SSLDM and MTLDM have the same structure. The latent space encoder $D_0$ and the conditional encoder are also identical. The distinction between the two lies in the need to train different decoders for different decomposed sub-images.

\begin{table}[h]
	\centering
		\caption{Parameters and their values.}
	\setlength{\tabcolsep}{4pt}  
	\begin{tabular}{m{4cm}|p{3cm}}  
		\hline
		Parameter Name & Value \\
		\hline
		Training Steps & 1.5M \\
		Sampling Steps & 1000 \\
		EMA Decay Rate & 0.998 \\
		EMA Update Step & 10 \\
		Base Learning Rate & le-4 \\
		Optimizer & Adam \\
		\hline
	\end{tabular}
	\label{trainingtab}
\end{table}

\begin{table}[htbp]
    \centering
    \caption{\textcolor{black}{Comparison of Model Sizes}}
    \label{tab:model_size}
    \begin{tabularx}{\linewidth}{>{\raggedright\arraybackslash}X>{\centering\arraybackslash}X}
        \toprule
        \textbf{Model} & \textbf{Parameters (M)} \\
        \midrule
        DGMR & 123.17 \\
        TRDMv1/v2 & 48.70 \\
        DiffCast & 47.07 \\
        MTLDM & 331.54 \\
        MTLDM(oa) & 331.54 \\
        \bottomrule
    \end{tabularx}
\end{table}

\begin{figure}[htbp]
	\centering
	\includegraphics[width=0.5\textwidth]{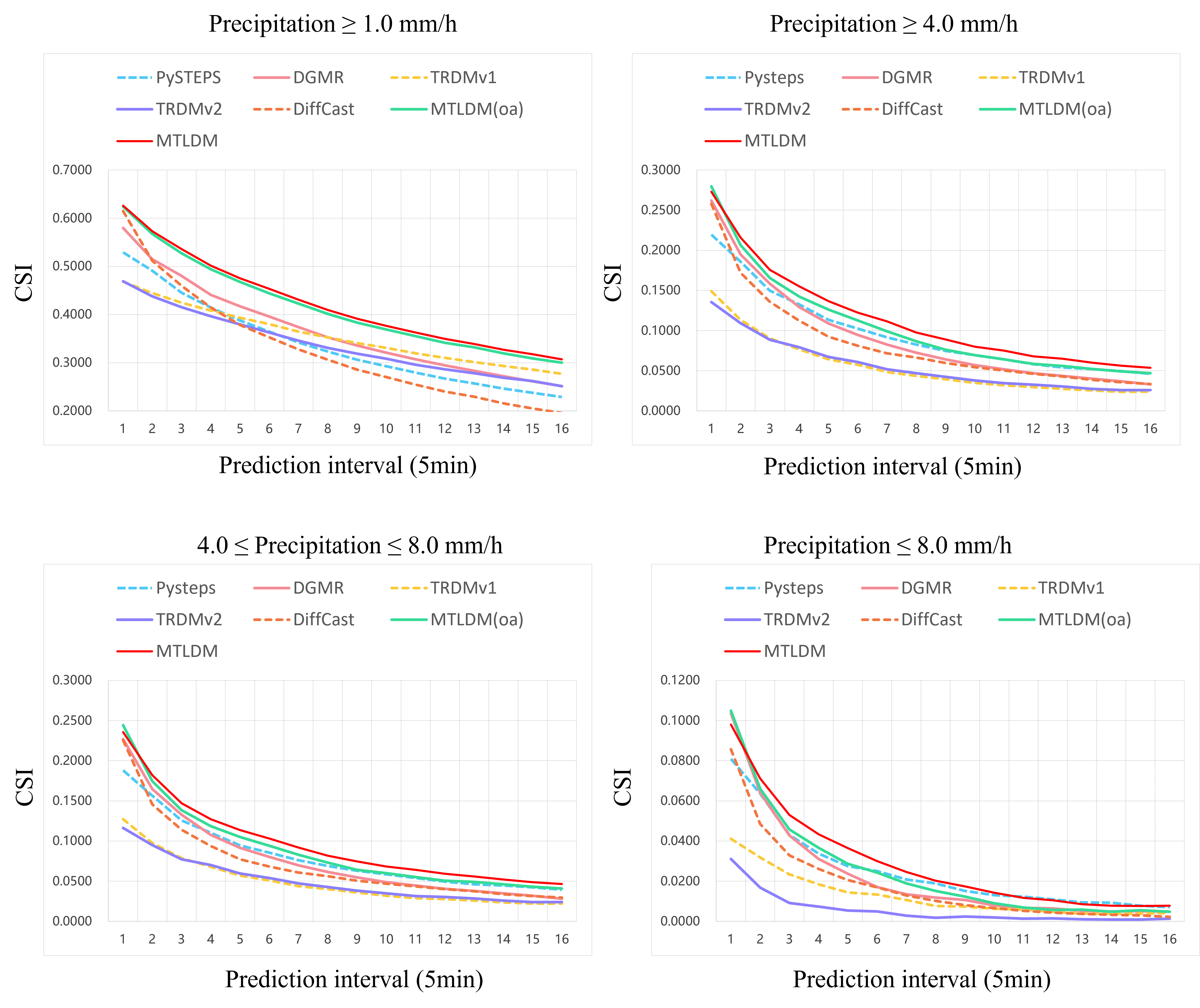}
	\caption{Comparison of CSI Across Prediction Intervals for Different Precipitation Intensity Thresholds on the MRMS dataset.}
	\label{fig6}
\end{figure}
\begin{figure}[htbp]
	\centering
	\includegraphics[width=0.5\textwidth]{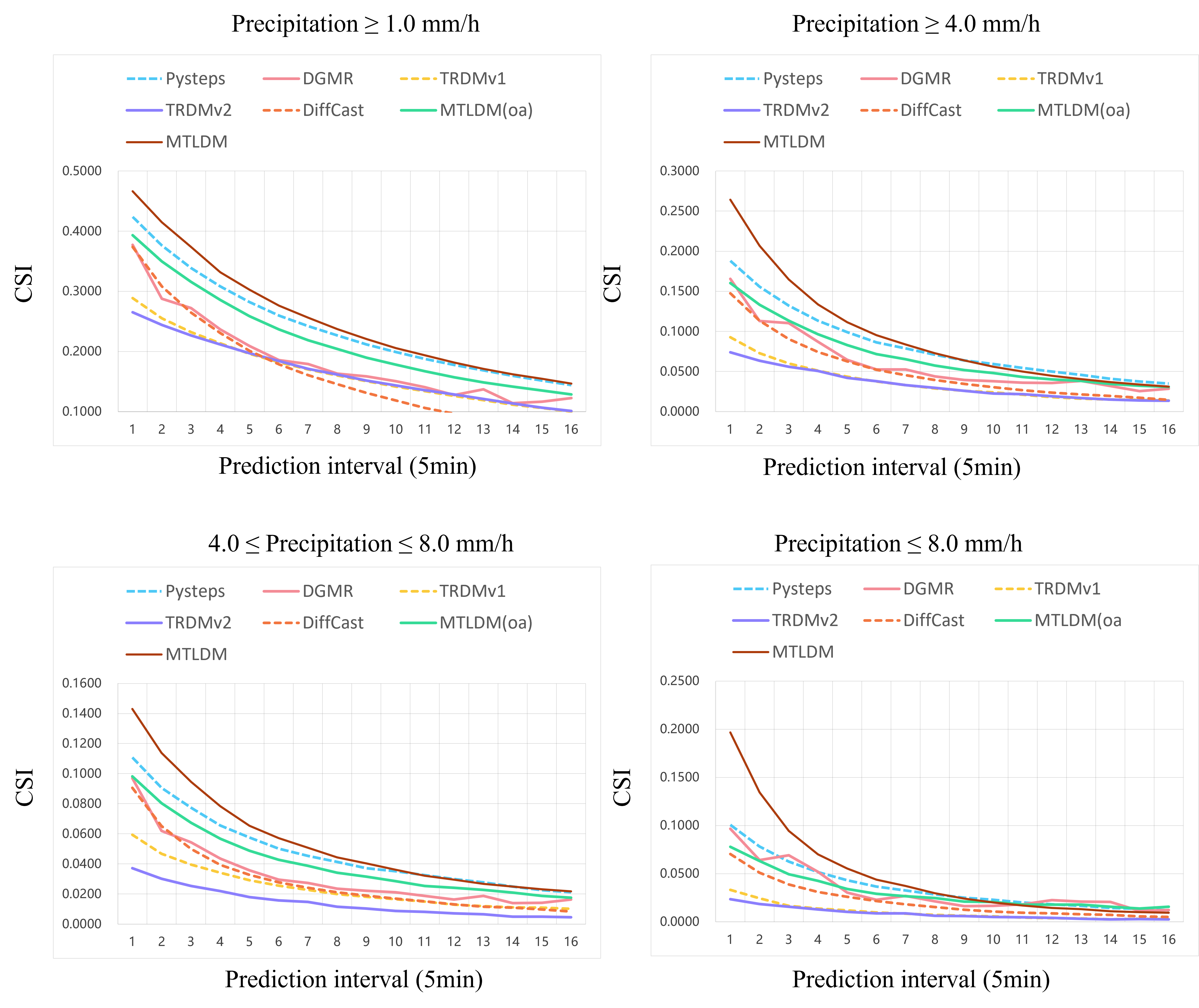}
	\caption{Comparison of CSI Across Prediction Intervals for Different Precipitation Intensity Thresholds on the SW dataset}
	\label{fig8}
\end{figure}

\begin{figure}[htbp]
	\centering
	\includegraphics[width=0.5\textwidth]{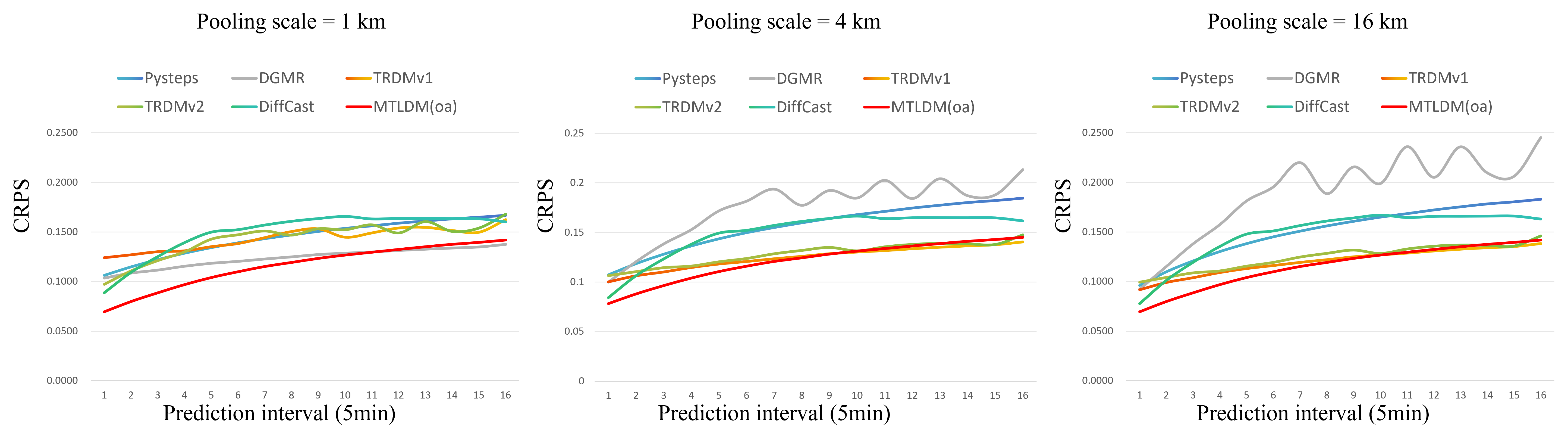}
	\caption{Comparison of CRPS Across Prediction Intervals for Different Precipitation Intensity Thresholds on the SW dataset}
	\label{fig9}
\end{figure}
\begin{figure}[htbp]
	\centering
	\includegraphics[width=0.5\textwidth]{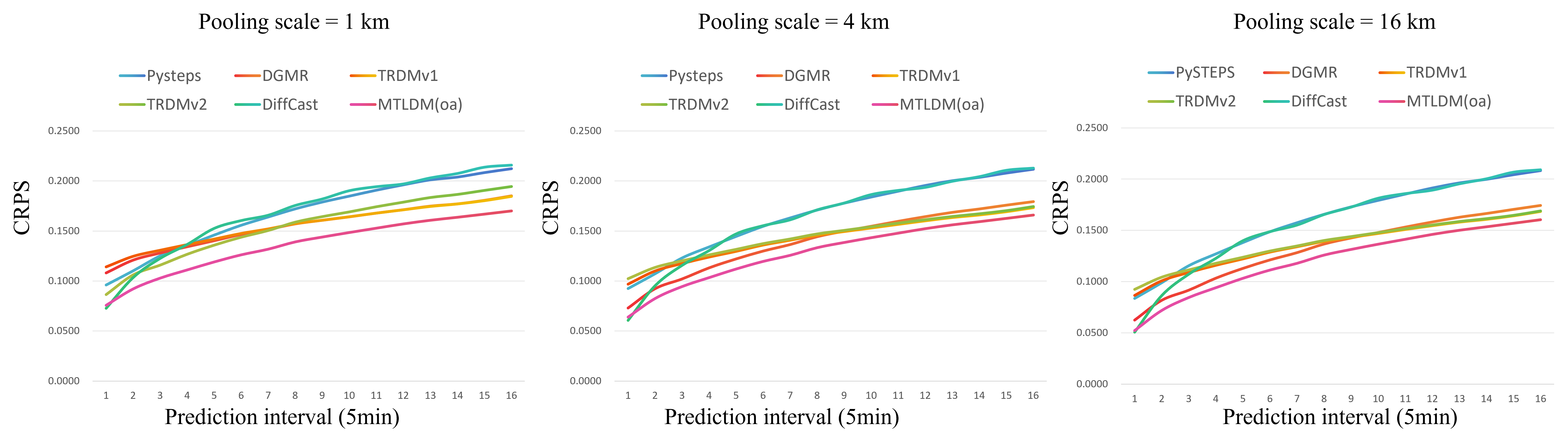}
	\caption{Comparison of CRPS Across Prediction Intervals for Different Precipitation Intensity Thresholds on the MRMS dataset on the MRMS dataset}
	\label{fig7}
\end{figure}

\subsection{Evaluation based on Metrics}
Each model predicts 16 future precipitation fields using the past four radar observation frames. To evaluate the performances of the models, we use common quantitative indicators (CRPS and CSI). All of the models except for PySTEPS have a $256\times256$ size input and output space and are evaluated by comparing the difference between the cropped ground truth in the testing set and the model's prediction results.

\begin{table*}[htbp]
	\setlength{\tabcolsep}{3pt}
	\centering 
	\caption{Comparison of precipitation CSI (↑) on the MRMS dataset. ↑ means the larger the better}
	\label{table2}
	\begin{tabular}{cccccccc|ccccccc}
		\toprule 
		No & PySTEPS & DGMR & TRDMv1 & TRDMv2 &\textcolor{black} {DiffCast}& MTLDM(oa) & MTLDM & PySTEPS & DGMR & TRDMv1 & TRDMv2  &\textcolor{black} {DiffCast}& MTLDM(oa) & MTLDM \\
	\midrule
		\multicolumn{8}{c|}{Precipitation $\geq$ 1.0 mm/h}  &\multicolumn{7}{c}{Precipitation $\geq$ 4.0 mm/h} \\
	\midrule
		1 & 0.5292 & 0.5798 & 0.4686 & 0.4692 &0.6147& 0.6243 & \textbf{0.6262} & 0.2194 & 0.2621 & 0.1489 & 0.1355 &0.2579& \textbf{0.2796 }& 0.2728 \\	 
		2 & 0.4912 & 0.5144 & 0.4448 & 0.4379 &0.5116& 0.5670 & \textbf{0.5728} & 0.1858 & 0.1949 & 0.1135 & 0.1094 &0.1718& 0.2066 &\textbf{ 0.2157} \\	 
		3 & 0.4455 & 0.4804 & 0.4249 & 0.4153 & 0.4593&0.5272 & \textbf{0.5359} & 0.1499 & 0.1585 & 0.0905 & 0.0887 &0.1358& 0.1654 & \textbf{0.1758 }\\	 
		4 & 0.4146 & 0.4407 & 0.4083 & 0.3963 &0.4148 &0.4943 & \textbf{0.5016 }& 0.1326 & 0.1292 & 0.0767 & 0.0792 &0.1122& 0.1422 &\textbf{ 0.1552} \\	 
		5 & 0.3875 & 0.4173 & 0.3931 & 0.3798 &0.3784& 0.4677 & \textbf{0.4752} & 0.1132 & 0.1090 & 0.0644 & 0.0672 &0.0925& 0.1261 &\textbf{ 0.1366 }\\	 
		6 & 0.3644 & 0.3958 & 0.3805 & 0.3629 &0.3525& 0.4439 & \textbf{0.4536} & 0.1027 & 0.0947 & 0.0576 & 0.0608 &0.0809& 0.1125 &\textbf{ 0.1224} \\	 
		7 & 0.3420 & 0.3739 & 0.3648 & 0.3461 & 0.3274&0.4229 & \textbf{0.4311 }& 0.0916 & 0.0827 & 0.0484 & 0.0520 &0.0718& 0.0990 & \textbf{0.1117 }\\	 
		8 & 0.3234 & 0.3530 & 0.3524 & 0.3310 & 0.3064&0.4015 & \textbf{0.4098} & 0.0826 & 0.0727 & 0.0437 & 0.0471 &0.0664& 0.0866 & \textbf{0.0978} \\		 
		9 & 0.3065 & 0.3358 & 0.3408 & 0.3191 & 0.2858&0.3831 & \textbf{0.3916} & 0.0748 & 0.0642 & 0.0396 & 0.0424 &0.0595& 0.0763 & \textbf{0.0888 }\\	 
		10 & 0.2930 & 0.3210 & 0.3306 & 0.3081 &0.2702 &0.3690 &\textbf{ 0.3766 }& 0.0697 & 0.0573 & 0.0351 & 0.0382 &0.0546& 0.0695 & \textbf{0.080} \\	 
		11 & 0.2794 & 0.3074 & 0.3199 & 0.2959 &0.2545 &0.3555 & \textbf{0.3628} & 0.0643 & 0.0519 & 0.0319 & 0.0345 &0.0503 &0.0642 & \textbf{0.0753} \\	 
		12 & 0.2668 & 0.2945 & 0.3101 & 0.2865 & 0.2402&0.3418 &\textbf{ 0.3494 }& 0.0584 & 0.0469 & 0.0298 & 0.0327 &0.0461 &0.0586 &\textbf{ 0.0682} \\	 
		13 & 0.2570 & 0.2830 & 0.3015 & 0.2784 & 0.2297&0.3318 & \textbf{0.3388 }& 0.0542 & 0.0437 & 0.0276 & 0.0306 & 0.0429&0.0562 & \textbf{0.0650} \\	 
		14 & 0.2467 & 0.2719 & 0.2933 & 0.2692 &0.2159& 0.3198 &\textbf{ 0.3271 }& 0.0522 & 0.0404 & 0.0256 & 0.0273 &0.0385 &0.0524 & \textbf{0.0601} \\	 
		15 & 0.2377 & 0.2617 & 0.2858 & 0.2618 &0.2053 &0.3092 &\textbf{ 0.3176} & 0.0494 & 0.0369 & 0.0238 & 0.0260 &0.0356& 0.0493 & \textbf{0.0564} \\	 
		16 & 0.2289 & 0.2512 & 0.2773 & 0.2515 &0.1960& 0.2999 &\textbf{ 0.3072} & 0.0462 & 0.0330 & 0.0242 & 0.0260 &0.0333 &0.0469 & \textbf{0.0537 }\\
		\hline
	\end{tabular}
\end{table*}

\begin{table*}[htbp]
	\setlength{\tabcolsep}{3pt}	
		\centering 
	\caption{Comparison of precipitation CSI (↑) on the MRMS dataset. ↑ means the larger the better}
	\label{table3}
		\begin{tabular}{cccccccc|ccccccc}
	\toprule
		No & PySTEPS & DGMR & TRDMv1 & TRDMv2 &\textcolor{black} {DiffCast} &MTLDM(oa) & MTLDM & PySTEPS & DGMR & TRDMv1 & TRDMv2&\textcolor{black} {DiffCast} & MTLDM(oa) & MTLDM \\
	\midrule
		\multicolumn{8}{c|}{Precipitation $> 4.0$ mm/h and $\leq 8.0$ mm/h}  &\multicolumn{6}{c}{Precipitation $\geq$ 8.0 mm/h} \\
	\midrule
	 1 & 0.1880 & 0.2271 & 0.1271 & 0.1163&0.2253& \textbf{0.2442 }& 0.2355 & 0.0810 & 0.1035 & 0.0411 & 0.0310 &0.0857& \textbf{0.1048 }& 0.0979 \\
	 2 & 0.1565 & 0.1647 & 0.0977 & 0.0950&0.1456 & 0.1744 &\textbf{ 0.1818} & 0.0636 & 0.0641 & 0.0318 & 0.0168 &0.0484& 0.0660 &\textbf{ 0.0711 }\\
	 3 & 0.1257 & 0.1329 & 0.0787 & 0.0771 &0.1141& 0.1386 &\textbf{ 0.1471 }& 0.0430 & 0.0428 & 0.0234 & 0.0092 &0.0327& 0.0457 &\textbf{ 0.0530} \\
	 4 & 0.1103 & 0.1077 & 0.0677 & 0.0700 &0.0938& 0.1187 & \textbf{0.1273 }& 0.0338 & 0.0311 & 0.0184 & 0.0073 &0.0261& 0.0366 & \textbf{0.0434} \\
	 5 & 0.0946 & 0.0916 & 0.0568 & 0.0597& 0.0773& 0.1051 &\textbf{ 0.1138 }& 0.0274 & 0.0238 & 0.0144 & 0.0054 &0.0205& 0.0286 &\textbf{ 0.0365} \\
	 6 & 0.0854 & 0.0801 & 0.0511 & 0.0539 &0.0682& 0.0942 &\textbf{ 0.1030} & 0.0250 & 0.0171 & 0.0133 & 0.0049 &0.0168& 0.0243 &\textbf{ 0.0300} \\
	 7 & 0.0760 & 0.0699 & 0.0438 & 0.0472 &0.0609& 0.0828 & \textbf{0.0919 }& 0.0208 & 0.0135 & 0.0107 & 0.0028 &0.0128& 0.0189 &\textbf{ 0.0246 }\\
	 8 & 0.0690 & 0.0615 & 0.0399 & 0.0429 &0.0563& 0.0731 &\textbf{ 0.0819 }& 0.0189 & 0.0119 & 0.0076 & 0.0018 &0.0102& 0.0152 &\textbf{ 0.0203} \\
	 9 & 0.0628 & 0.0546 & 0.0360 & 0.0384 &0.0506& 0.0642 & \textbf{0.0747} & 0.0151 & 0.0106 & 0.0073 & 0.0025 &0.0081& 0.0122 & \textbf{0.0175 }\\
	 10 & 0.0586 & 0.0486 & 0.0320 & 0.0350& 0.0467& 0.0599 & \textbf{0.0681 }& 0.0131 & 0.0080 & 0.0063 & 0.0019 &0.0065& 0.0091 &\textbf{ 0.0143 }\\
	 11 & 0.0540 & 0.0444 & 0.0290 & 0.0313 &0.0434& 0.0549 & \textbf{0.0641} & \textbf{0.0122 }& 0.0067 & 0.0060 &0.0014 & 0.0052& 0.0070 & 0.0117 \\
	 12 & 0.0497 & 0.0404 & 0.0276 & 0.0302& 0.0400& 0.0506 &\textbf{ 0.0591} &\textbf{ 0.0109} & 0.0064 & 0.0048 & 0.0015 &0.0037& 0.0056 & 0.0105 \\
	 13 & 0.0462 & 0.0379 & 0.0259 & 0.0285& 0.0373& 0.0493 &\textbf{ 0.0560 }&\textbf{ 0.0095} & 0.0052 & 0.0038 & 0.0011 &0.0037& 0.0059 & 0.0086 \\
	 14 & 0.0448 & 0.0348 & 0.0238 & 0.0257&0.0337 & 0.0462 & \textbf{0.0520} & \textbf{0.0093} & 0.0050 & 0.0038 & 0.0008 &0.0032& 0.0048 & 0.0079 \\
	 15 & 0.0422 & 0.0318 & 0.0221 & 0.0241 & 0.0314& 0.0432 & \textbf{0.0487 }&\textbf{ 0.0079 }& 0.0051 & 0.0039 & 0.0009 &0.0029& 0.0055 & 0.0076 \\
	 16 & 0.0393 & 0.0282 & 0.0222 & 0.0239 &0.0295& 0.0407 &\textbf{ 0.0467} & 0.0071 & 0.0048 & 0.0041 & 0.0013 &0.0022& 0.0048 & \textbf{0.0078} \\
	 
		\hline
	\end{tabular}
\end{table*}
\begin{table*}[htbp]
		\centering 
		
		\setlength{\tabcolsep}{3pt}  
		\caption{Comparison of precipitation CSI (↑) on the SW dataset. ↑ means the larger the better}
		\label{table4}
		\begin{tabular}{cccccccc|ccccccc}
	\toprule
		No & PySTEPS & DGMR & TRDMv1 & TRDMv2&\textcolor{black} {DiffCast} & MTLDM(oa) & MTLDM & PySTEPS & DGMR & TRDMv1 & TRDMv2 &\textcolor{black} {DiffCast }& MTLDM(oa) & MTLDM \\
		\midrule
		\multicolumn{8}{c|}{Precipitation $\geq$ 1.0 mm/h}  &\multicolumn{6}{c}{Precipitation $\geq$ 4.0 mm/h} \\
		\midrule
			1 & 0.4235 & 0.3773 & 0.2889 & 0.2655&0.3735 & 0.3934 &\textbf{ 0.4663} & 0.1882 & 0.1657 & 0.0928 & 0.0740 &0.1475& 0.1602 &\textbf{ 0.2642} \\
			2 & 0.3765 & 0.2878 & 0.2550 & 0.2443 &0.3084& 0.3497 & \textbf{0.4148} & 0.1557 & 0.1132 & 0.0729 & 0.0634 &0.1133& 0.1331 & \textbf{0.2067} \\
			3 & 0.3386 & 0.2724 & 0.2322 & 0.2267 &0.2645& 0.3155 & \textbf{0.3737} & 0.1320 & 0.1105 & 0.0602 & 0.0559 &0.0904& 0.1134 & \textbf{0.1649} \\
			4 & 0.3084 & 0.2368 & 0.2137 & 0.2117&0.2307 & 0.2855 & \textbf{0.3318} & 0.1135 & 0.0871 & 0.0512 & 0.0507 &0.0724& 0.0965 & \textbf{0.1337} \\
			5 & 0.2821 & 0.2090 & 0.1962 & 0.1971 & 0.2014& 0.2588 &\textbf{ 0.3029} & 0.0993 & 0.0647 & 0.0438 & 0.0423 & 0.0627&0.0829 & \textbf{0.111}4 \\
			6 & 0.2596 & 0.1859 & 0.1821 & 0.1840& 0.1784& 0.2365 & \textbf{0.2761 }& 0.0865 & 0.0527 & 0.0375 & 0.0379 &0.0522& 0.0717 &\textbf{ 0.0954 }\\
			7 & 0.2420 & 0.1791 & 0.1701 & 0.1711 & 0.1607& 0.2188 & \textbf{0.2564 }& 0.0787 & 0.0528 & 0.0330 & 0.0331&0.0454 & 0.0655 & \textbf{0.0836} \\
			8 & 0.2265 & 0.1634 & 0.1602 & 0.1614 & 0.1454& 0.2041 & \textbf{0.2371 }& 0.0710 & 0.0441 & 0.0291 & 0.0296&0.0395 & 0.0576 &\textbf{ 0.0728} \\
			9 & 0.2116 & 0.1586 & 0.1508 & 0.1518 &0.1310& 0.1898 & \textbf{0.2206 }& \textbf{0.0639} & 0.0393 & 0.0261 & 0.0259 &0.0347 &0.0518 & \textbf{0.0639} \\
			10 & 0.1991 & 0.1507 & 0.1414 & 0.1437& 0.1183& 0.1781 &\textbf{ 0.2055 }&\textbf{ 0.0593} & 0.0379 & 0.0240 & 0.0226 & 0.0305&0.0481 & 0.0561 \\
			11 & 0.1879 & 0.1408 & 0.1342 & 0.1356& 0.1062& 0.1672 & \textbf{0.1934} & \textbf{0.0544} & 0.0362 & 0.0210 & 0.0218 &0.0268& 0.0433 & 0.0501 \\
			12 & 0.1778 & 0.1279 & 0.1260 & 0.1286&0.0967 & 0.1573 & \textbf{0.1816} &\textbf{ 0.0499 }& 0.0359 & 0.0183 & 0.0194 &0.0238& 0.0403 & 0.0446 \\
			13 & 0.1685 & 0.1371 & 0.1191 & 0.1210 &0.0876& 0.1489 & \textbf{0.1712 }&\textbf{ 0.0458} & 0.0382 & 0.0164 & 0.0170 &0.0214 &0.0383 & 0.0404 \\
			14 & 0.1594 & 0.1142 & 0.1117 & 0.1136&0.0786 & 0.1418 & \textbf{0.1619 }& \textbf{0.0409} & 0.0320 & 0.0151 & 0.0152 &0.0197& 0.0346 & 0.0367 \\
			15 & 0.1519 & 0.1164 & 0.1059 & 0.1069 &0.0710& 0.1350 & \textbf{0.1546 }&\textbf{ 0.0376 }& 0.0258 & 0.0149 & 0.0141 & 0.0172&0.0323 & 0.0339 \\
			16 & 0.1444 & 0.1227 & 0.1001 & 0.1014& 0.0634& 0.1288 & \textbf{0.1464 }& \textbf{0.0349} & 0.0288 & 0.0140 & 0.0134 &0.0148& 0.0310 & 0.0311 \\
			\bottomrule
		\end{tabular}
	\end{table*}
	
	\begin{table*}[htbp]	
		\centering 
		\setlength{\tabcolsep}{3pt}
		\caption{Comparison of precipitation CSI (↑) on the SW dataset. ↑ means the larger the better}
		\label{table5}
		\begin{tabular}{cccccccc|ccccccc}
			\toprule
			No & PySTEPS & DGMR & TRDMv1 & TRDMv2 &\textcolor{black} {DiffCast}& MTLDM(oa) & MTLDM & PySTEPS & DGMR & TRDMv1 & TRDMv2 &\textcolor{black} {DiffCast}& MTLDM(oa) & MTLDM \\
			\midrule
			\multicolumn{8}{c|}{Precipitation $> 4.0$ mm/h and $\leq 8.0$ mm/h}  &\multicolumn{6}{c}{Precipitation $\geq$ 8.0 mm/h} \\
			\midrule
			1 & 0.1108 & 0.0970 & 0.0594 & 0.0372 & 0.0906&0.0983 &\textbf{ 0.1430 }& 0.1008 & 0.0966 & 0.0332 & 0.0236 &  0.0704&0.0780 & \textbf{0.1966} \\
			2 & 0.0906 & 0.0621 & 0.0468 & 0.0301 &0.0650 &0.0805 & \textbf{0.1137 }& 0.0782 & 0.0641 & 0.0245 & 0.0186 & 0.0509 &0.0631 &\textbf{ 0.1344} \\
			3 & 0.0773 & 0.0544 & 0.0396 & 0.0255 & 0.0500&0.0674 & \textbf{0.0947} & 0.0624 & 0.0693 & 0.0167 & 0.0155 &0.0387 & 0.0495 &\textbf{ 0.0944 }\\
			4 & 0.0656 & 0.0435 & 0.0341 & 0.0219 &0.0393& 0.0568 & \textbf{0.0783 }& 0.0515 & 0.0518 & 0.0139 & 0.0128 & 0.3100 &0.0427 &\textbf{ 0.0701} \\
			5 & 0.0577 & 0.0357 & 0.0292 & 0.0179 &0.0328& 0.0487 & \textbf{0.0654 }& 0.0430 & 0.0303 & 0.0118 & 0.0102 &0.2612 &0.0341 & \textbf{0.0555} \\
			6 & 0.0501 & 0.0295 & 0.0256 & 0.0158 &0.0277& 0.0428 & \textbf{0.0572 }& 0.0368 & 0.0230 & 0.0098 & 0.0089 & 0.0215&0.0290 & \textbf{0.0437} \\
			7 & 0.0454 & 0.0273 & 0.0228 & 0.0149 &0.0241& 0.0387 & \textbf{0.0508 }& 0.0325 & 0.0268 & 0.0085 & 0.0088 &0.0183& 0.0266 & \textbf{0.0374} \\
			8 & 0.0414 & 0.0235 & 0.0200 & 0.0116 & 0.021&0.0341 &\textbf{ 0.0444 }& 0.0285 & 0.0212 & 0.0071 & 0.0064 & 0.0152& 0.0247 & \textbf{0.0298 }\\
			9 & 0.0373 & 0.0221 & 0.0181 & 0.0104 & 0.0189&0.0316 & \textbf{0.0403} & \textbf{0.0250 }& 0.0162 & 0.0061 & 0.0125&0.0012 & 0.0210 & 0.0242 \\
			10 & 0.0351 & 0.0212 & 0.0166 & 0.0087 &0.0169& 0.0285 & \textbf{0.0363 }& \textbf{0.0228 }& 0.0167 & 0.0057 & 0.0052 & 0.0107& 0.0207 & 0.0199 \\
			11 & \textbf{0.0326} & 0.0188 & 0.0149 & 0.0082 & 0.0150&0.0254 & 0.0322 & \textbf{0.0199 }& 0.0181 & 0.0045 & 0.0047&0.0094 & 0.0178 & 0.0169 \\
			12 &\textbf{ 0.0299} & 0.0164 & 0.0129 & 0.0072 &0.0150& 0.0242 & 0.0296 & 0.0183 & \textbf{0.0225} & 0.0037 & 0.0040 & 0.0087&0.0177 & 0.0145 \\
			13 & \textbf{0.0277} & 0.0187 & 0.0119 & 0.0065 &0.0131& 0.0227 & 0.0268 & 0.0165 & \textbf{0.0210 }& 0.0034 & 0.0033 & 0.0077&0.0180 & 0.0133 \\
			14 & 0.0249 & 0.0139 & 0.0111  & 0.0049 & 0.0209& 0.0116&\textbf{ 0.0250}  & 0.0146  & \textbf{0.0207} & 0.0027 & 0.0026 &0.0070& 0.0158 & 0.0108\\
			15 & 0.0249 & 0.0142 & 0.0109 & 0.0050 &0.0097& 0.0188 & \textbf{0.0231 }& 0.0136 & 0.0114 & 0.0031 & 0.0028 &0.0058& \textbf{0.0138} & 0.0102 \\
			16 & 0.0212 & 0.0165 & 0.0102 & 0.0046 &0.0084& 0.0175 &\textbf{ 0.0219 }& 0.0120 & 0.0123 & 0.0029 & 0.0024 &0.0049& \textbf{0.0157 }& 0.0095 \\
			\bottomrule
		\end{tabular}
	\end{table*}
	
	\begin{figure*}
		\centering
		\includegraphics[width=0.9\textwidth]{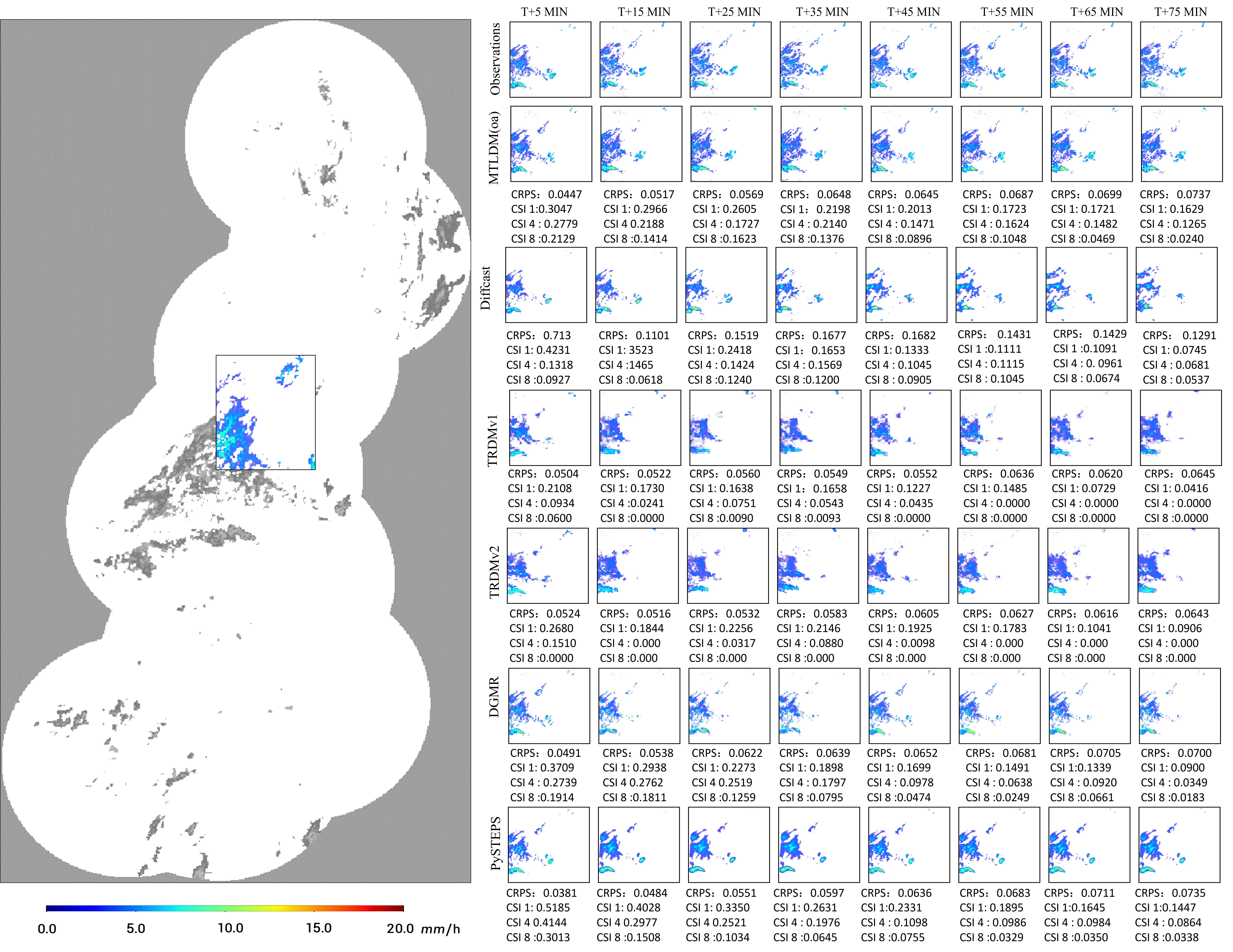}
		\caption{Prediction performance and visualization evaluation of the MTLDM and other models for a precipitation event that occurred on August 01, 2021, at 18:35:00 in the SW dataset.}
		\label{fig-1}
	\end{figure*}
	\begin{figure*}
		\centering
		\includegraphics[width=0.9\textwidth]{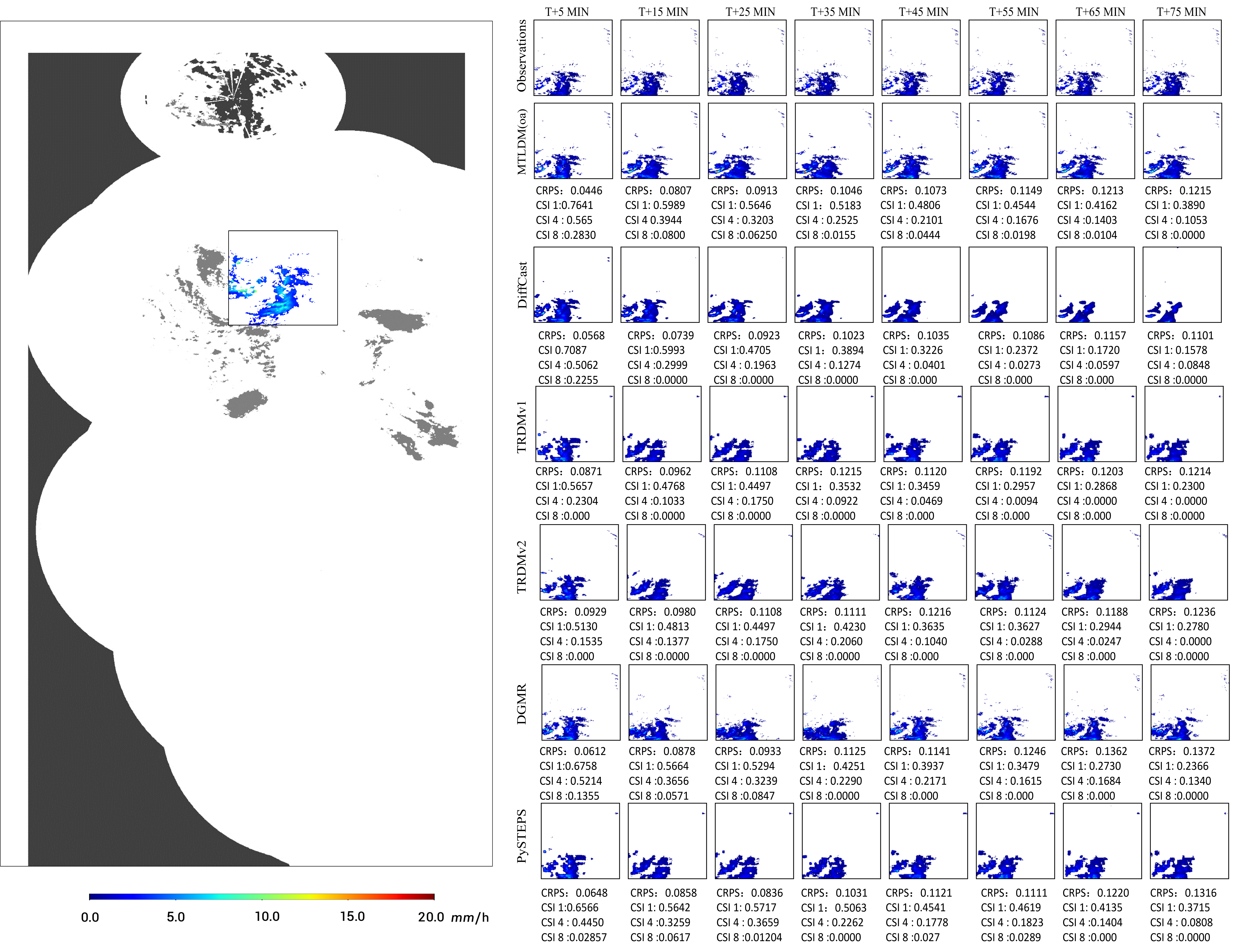}
		\caption{Prediction performance and visualization evaluation of the MTLDM model for a precipitation event that occurred on April 10, 2022, at 15:16:00 in the MRMS dataset}
		\label{fig-2}
	\end{figure*}
	
\textbf{CSI Analysis}
According to Ravuri et al. \cite{ravuri2021skilful}, we evaluate our model’s predictive ability for precipitation intervals of $\geq$ 1 mm/h, and  $\geq$ 8.0 mm/h, and we also provide its performance for the 4–8 mm/h interval. Fig. \ref{fig6} shows the CSI results of sevaral prediction models on the MRMS dataset. The detailed CSI indicator data of the prediction results of each model on the MRMS dataset are listed in Table\ref{table2} and Table\ref{table3}. For the precipitation interval of $\geq$ 1 mm/h, the performance of the diffusion model is superior to those of the non-diffusion models, and our proposed MTLDM far outperforms the other methods, including the MTLDM(oa). \textcolor{black}{On average, its CSI is 13\% higher than DGMR, 23\% higher than PySTEPS, and 26.5\% higher than Diffcast}. For the $\geq$ 4 mm/h region, the CSI of the MTLDM is about 20\% higher on average than those of the DGMR and PySTEPS and also outperforms DiffCast. For the heavy precipitation intervals of 4-8 and $\geq$8 mm/h, the MTLDM also significantly outperforms the other four methods. Fig.\ref{fig8} shows the results of the performance comparison of the six precipitation forecast models for forecasting precipitation events with $\geq$1, 4, 4-8, and $\geq$8 mm/h in the SW datasets according to the CSI, and the detailed data of the models are listed in Table \ref{table4} and Table \ref{table5} . The MTLDM model has the best overall performance over the entire forecast period, followed by MTLDM (oa) with similar performance. \textcolor{black}{DiffCast performed better than TRDM models but not as well as MTLDM}. PySTEPS and the DGMR performed similarly, with better results in short-term forecasts, but their performances rapidly degraded over time. At the beginning of the forecast, for the MRMS dataset, the maximum CSI differences between the MTLDM and the other models were 0.16, 0.14, 0.12, and 0.05 for the $\geq$1, $\geq$4, 4–8, and $\geq$8 mm/h scenarios, respectively. For the SW dataset, these differences were approximately 0.11, 0.15, 0.15, and 0.17, respectively. Although the gap has narrowed over time, the MTLDM still maintained the top position.

The traditional PySTEPS model relies on the robustness of its physical principles to show unique advantages in handling rare events. Especially when predicting extreme precipitation events with intensities of greater than 8 mm/h, its performance exceeds those of most of the deep learning models. This phenomenon highlights the advantages of traditional methods based on physical principles in the face of data sparsity and also reveals the limitations of purely data-driven methods. However, the MTLDM model developed in this study performs well under this challenge. Even under high-intensity precipitation of $\geq$ 8 mm/h, the MTLDM still clearly maintains the top position. This result proves the MTLDM's excellent ability to handle extreme events. The staged decoding idea we propose effectively overcomes the dilemma of a lack of data, allowing the MTLDM to maintain its significant advantages during scarce precipitation events with $\geq$8 mm/h.
	
\begin{table*}[htbp]
 \centering  
	\caption{Comparison of precipitation CRPS (↓) on the SW dataset. ↓ means the smaller the better}
	\label{table6}
\resizebox{\textwidth}{!}{%
	\begin{tabular}{lcccccc|cccccc|cccccc}
		\toprule
		\multirow{2}{*}{\textbf{No}} & \multicolumn{6}{c|}{\textbf{Pooling scale = 1 km}} & \multicolumn{6}{c|}{\textbf{Pooling scale = 4 km}} & \multicolumn{6}{c}{\textbf{Pooling scale = 16 km}} \\
		\cmidrule(lr){2-7} \cmidrule(lr){8-13} \cmidrule(lr){13-19}
		& \textbf{PySTEPS} & \textbf{DGMR} & \textbf{TRDMv1} & \textbf{TRDMv2} & \textcolor{black} {\textbf{DiffCast}}& \textbf{MTLDM(oa)} & \textbf{PySTEPS} & \textbf{DGMR} & \textbf{TRDMv1} & \textbf{TRDMv2} & \textcolor{black} {\textbf{DiffCast}}&\textbf{MTLDM(oa)} & \textbf{PySTEPS} & \textbf{DGMR} & \textbf{TRDMv1} & \textbf{TRDMv2} &\textcolor{black} {\textbf{DiffCast}} & \textbf{MTLDM(oa)} \\
		\midrule
 1 & 0.1062 & 0.0973 & 0.1036 & 0.1242 &0.0887&\textbf{0.0842}& 0.1073 & 0.0999 & 0.1001 & 0.1063& 0.0843&  \textbf{0.0783} & 0.0962 & 0.0924 & 0.0918 & 0.0995 & 0.0776&\textbf{ 0.0694 }\\
 2 & 0.1148 & 0.1108 & 0.1086 & 0.1271&0.1092 &\textbf{ 0.0922} & 0.1185 & 0.1205 & 0.1062 &  0.1104& 0.1064& \textbf{0.0880} & 0.1099 & 0.1153 & 0.0992 & 0.1042 &0.1011& \textbf{0.0798 }\\
 3 & 0.1221 & 0.1212 & 0.1116 & 0.1302& 0.1251&\textbf{ 0.1003 }& 0.1281 & 0.1385 & 0.1101 &  0.1144 & 0.1233& \textbf{0.0965} & 0.1209 & 0.1381 & 0.1039 & 0.1088 &0.1194& \textbf{0.0886} \\
 4 & 0.1286 & 0.1299 & 0.1155 & 0.1311&0.1392 & \textbf{0.1072 }& 0.1364 & 0.1528 & 0.1146  & 0.1160 &0.1381&  \textbf{0.1040 }& 0.1303 & 0.1575 & 0.1092 & 0.1108 &0.1353 &\textbf{0.0969} \\
 5 & 0.1342 & 0.1428 & 0.1185 & 0.1353 &0.1497 & \textbf{0.1123 }& 0.1436 & 0.1719 & 0.1181 &  0.1204 &0.1494&\textbf{ 0.1106} & 0.1383 & 0.1814 & 0.1132 & 0.1156 &0.1478 &\textbf{0.1040 }\\
 6 & 0.1391 & 0.1472 & 0.1204 & 0.1384 &0.1522 &\textbf{ 0.1175 }& 0.1498 & 0.1816 & 0.1206 & 0.1237 &0.1522 &\textbf{0.1160} & 0.1451 & 0.1954 & 0.1162 & 0.1194 &0.1509& \textbf{0.1099 }\\
 7 & 0.1432 & 0.1509 & 0.1228 & 0.1444 & 0.1569&\textbf{ 0.1213 }& 0.1550 & 0.1938 & 0.1234 & 0.1285 & 0.1571& \textbf{0.1208} & 0.1509 & 0.2199 & 0.1194 & 0.1248 & 0.1564&\textbf{0.1152} \\
 8 & 0.1470 & 0.1468 & 0.1249 & 0.1505&0.1532 &\textbf{ 0.1245} & 0.1597 & 0.1774 & 0.1258 & 0.1319 &0.1612 &\textbf{ 0.1244} & 0.1561 & 0.1887 & 0.1221 & 0.1285 &0.1609 &\textbf{0.1193} \\
 9 & 0.1505 & 0.1533 & \textbf{0.1273 }&0.1634  &0.1607 & 0.1279 & 0.1641 & 0.1924 & 0.1284 & 0.1349 &0.1641& \textbf{0.1282}&  0.1608 & 0.2156 & 0.1250 & 0.1318 &0.1640& \textbf{0.1234} \\
 10 & 0.1536 & 0.1522 & \textbf{0.1287} & 0.1448&0.1655 & 0.1316 & 0.1680 & 0.1848 & \textbf{0.1301 }& 0.1314 & 0.1665& 0.1312 & 0.1650 & 0.1989 & 0.1270 & 0.1285 &0.1669&\textbf{ 0.1267} \\
 11 & 0.1562 & 0.1572 &\textbf{ 0.1299} & 0.1491& 0.1630 & 0.1335 & 0.1712 & 0.2027 &\textbf{ 0.1316} & 0.1355&0.1640 & 0.1338 & 0.1685 & 0.2360 & \textbf{0.1288 }& 0.1329 &0.1645& 0.1296 \\
 12 & 0.1588 & 0.1490 &\textbf{ 0.1314} & 0.1540 &0.1636 & 0.1361 & 0.1746 & 0.1843 & \textbf{0.1334} & 0.1380 & 0.1649& 0.1364 & 0.1722 & 0.2051 & \textbf{0.1309} & 0.1358 &0.1656& 0.1325 \\
 13 & 0.1610 & 0.1604 &\textbf{ 0.1327} & 0.1547 &0.1635 & 0.1378 & 0.1775 & 0.2043 &\textbf{ 0.1350 }& 0.1391& 0.1649& 0.1388 & 0.1754 & 0.2359 &\textbf{ 0.1328} & 0.1369 & 0.1657&0.1352 \\
 14 & 0.1632 & 0.1506 & \textbf{0.1338} & 0.1515& 0.1634& 0.1399 & 0.1802 & 0.1872 &\textbf{ 0.1363} & 0.1386 &0.1649  & 0.1412 & 0.1783 & 0.2096 &\textbf{ 0.1343 }& 0.1366 & 0.1659&0.1376 \\
 15 & 0.1648 & 0.1539 & \textbf{0.1349} & 0.1498& 0.1632& 0.1419 & 0.1822 & 0.1880 & \textbf{0.1375 }& 0.1379 &0.1648 & 0.1429 & 0.1804 & 0.2062 &\textbf{ 0.1354 }& 0.1359 & 0.1660&0.1396 \\
 16 & 0.1668 & 0.1679 & \textbf{0.1375 }& 0.1624 &0.1600 & 0.1440 & 0.1846 & 0.2135 & \textbf{0.1403} & 0.1477 & 0.1661& 0.1452 & 0.1830 & 0.2452 & \textbf{0.1384 }& 0.1461 & 0.1629&0.1420 \\
 	\bottomrule
	\end{tabular}%
}
\end{table*}

 \begin{table*}[!t] 
 	\centering
 	\caption{Comparison of precipitation CRPS (↓) on the MRMS dataset. ↓ means the smaller the better}
 		\label{table7}
 	\resizebox{\textwidth}{!}{%
 	\begin{tabular}{lcccccc|cccccc|cccccc}
 		\toprule
 		\multirow{2}{*}{\textbf{No}} & \multicolumn{6}{c|}{\textbf{Pooling scale = 1 km}} & \multicolumn{6}{c|}{\textbf{Pooling scale = 4 km}} & \multicolumn{6}{c}{\textbf{Pooling scale = 16 km}} \\
 		\cmidrule(lr){2-7} \cmidrule(lr){8-13} \cmidrule(lr){13-19}
 		& \textbf{PySTEPS} & \textbf{DGMR} & \textbf{TRDMv1} & \textbf{TRDMv2} &\textcolor{black}{\textbf{DiffCast}}& \textbf{MTLDM(oa)} & \textbf{PySTEPS} & \textbf{DGMR} & \textbf{TRDMv1} & \textbf{TRDMv2} &\textcolor{black}{\textbf{ DiffCast}}&\textbf{MTLDM(oa)} & \textbf{PySTEPS} & \textbf{DGMR} & \textbf{TRDMv1} & \textbf{TRDMv2} &\textcolor{black}{\textbf{ DiffCast}}&\textbf{MTLDM(oa)} \\
 		\midrule
 1 & 0.0960 & 0.0866 & 0.1081 & 0.1141 &\textbf{ 0.0725}&0.0758& 0.0923 & 0.0731 & 0.0968 & 0.1024 &\textbf{0.0605}&0.0638& 0.0835 & 0.0624 & 0.0865 & 0.0924 &\textbf{0.0508}& 0.0524 \\
 2 & 0.1101 & 0.1063 & 0.1209 & 0.1247 &0.1033&\textbf{ 0.0921 }& 0.1071 & 0.0924 & 0.1100 & 0.1137 &0.0951&\textbf{ 0.0824} & 0.0988 & 0.0817 & 0.1007 & 0.1043 & 0.8644&\textbf{0.0719 }\\
 3 & 0.1251 & 0.1160 & 0.1280 & 0.1308 &0.1225& \textbf{0.1029 }& 0.1230 & 0.1022 & 0.1174 & 0.1203 &0.1155&\textbf{ 0.0944 }& 0.1155 & 0.0916 & 0.1088 & 0.1114 &0.1073&\textbf{ 0.0846 }\\
 4 & 0.1354 & 0.1269 & 0.1342 & 0.1364& 0.1365&\textbf{ 0.1111 }& 0.1340 & 0.1133 & 0.1239 & 0.1262 &0.1303& \textbf{0.1034 }& 0.1270 & 0.1033 & 0.1158 & 0.1179 & 0.1227& \textbf{0.0941 }\\
 5 & 0.1459 & 0.1360 & 0.1402 & 0.1419 &0.1526& \textbf{0.1191 }& 0.1445 & 0.1222 & 0.1296 & 0.1317 &0.1470&\textbf{ 0.1120} & 0.1381 & 0.1128 & 0.1221 & 0.1237 &0.1399&\textbf{ 0.1033 }\\
 6 & 0.1557 & 0.1440 & 0.1465 & 0.1476 &0.1605& \textbf{0.1262 }& 0.1546 & 0.1300 & 0.1359 & 0.1375 &0.1552& \textbf{0.1196 }& 0.1487 & 0.1211 & 0.1287 & 0.1298 &0.1487& \textbf{0.1113} \\
 7 & 0.1640 & 0.1506 & 0.1515 & 0.1521 &0.1658& \textbf{0.1318} & 0.1630 & 0.1366 & 0.1408 & 0.1421 &0.1611& \textbf{0.1257} & 0.1575 & 0.1282 & 0.1339 & 0.1347 &0.1552&\textbf{ 0.1179} \\
 8 & 0.1721 & 0.1590 & 0.1571 &0.1573 &0.1756& \textbf{0.1391 }& 0.1708 & 0.1444 & 0.1457 & 0.1471 &0.1711& \textbf{0.1334} & 0.1657 & 0.1366 & 0.1392 & 0.1401 &0.1654&\textbf{ 0.1259} \\
 9 & 0.1789 & 0.1645 & 0.1606 & 0.1608 &0.1820 &\textbf{0.1440} & 0.1777 & 0.1500 & 0.1494 & 0.1507 &0.1779& \textbf{0.1386 }& 0.1729 & 0.1425 & 0.1431 & 0.1439 &0.1727&\textbf{ 0.1314} \\
 10 & 0.1848 & 0.1692 & 0.1641 & 0.1642 &0.1903& \textbf{0.1485} & 0.1838 & 0.1549 & 0.1530 & 0.1543 &0.1864& \textbf{0.1434} & 0.1793 & 0.1478 & 0.1469 & 0.1477 & 0.1814&\textbf{0.1365} \\
 11 & 0.1908 & 0.1744 & 0.1678 & 0.1679 &0.1943& \textbf{0.1528 }& 0.1896 & 0.1598 & 0.1566 & 0.1579 &0.1905& \textbf{0.1479 }& 0.1854 & 0.1531 & 0.1507 & 0.1516 &0.1858& \textbf{0.1413} \\
 12 & 0.1961 & 0.1790 & 0.1712  &0.1711& 0.1971& \textbf{0.1571 }& 0.1954 & 0.1645 & 0.1602 & 0.1613 &0.1953& \textbf{0.1524} & 0.1914 & 0.1581 & 0.1544 & 0.1551 & 0.1892&\textbf{0.1461} \\
 13 & 0.2012 & 0.1836 & 0.1748  &0.1745 &0.2032& \textbf{0.1607 }& 0.2001 & 0.1687 & 0.1634 & 0.1645 &0.1997&\textbf{ 0.1562} & 0.1963 & 0.1627 & 0.1579 & 0.1586 &0.1954 &\textbf{0.1501 }\\
 14 & 0.2039 & 0.1867 & 0.1770 & 0.1770 &0.2075& \textbf{0.1637 }& 0.2035 & 0.1722 & 0.1660 & 0.1672 &0.2042& \textbf{0.1594 }& 0.1999 & 0.1663 & 0.1606 & 0.1613 &0.2003&\textbf{ 0.1535 }\\
 15 & 0.2084 & 0.1907 & 0.1806 & 0.1803 &0.2139& \textbf{0.1668} & 0.2078 & 0.1760 & 0.1693 & 0.1703 &0.2106& \textbf{0.1627} & 0.2043 & 0.1703 & 0.1641 & 0.1646 &0.2068&\textbf{ 0.1569 }\\
 16 & 0.2123 & 0.1946 & 0.1850 & 0.1846 &0.2159&\textbf{ 0.1700 }& 0.2116 & 0.1795 & 0.1733 & 0.1744 &0.2128&\textbf{ 0.1660} & 0.2084 & 0.1742 & 0.1683 & 0.1689 &0.2091& \textbf{0.1604 }\\
 		\bottomrule
 	\end{tabular}
 } 
 \end{table*}
 
 \textbf{CRPS Analysis}
 The CRPS describes the overall precipitation distribution, so we use the overall forecast radar image to evaluate it, that is, we calculate the index of the MTLDM (oa) (without any ambiguity, we will refer to it as the MTLDM). Fig.\ref{fig9} and Fig.\ref{fig7} show the CRPS values of the radar images at scales of 1, 4, and 16, and the detailed data of the comparative experiments are shown in Tables \ref{table6} and \ref{table7}. The image scaling is achieved through pooling operations. The experimental results consistently demonstrate the superiority of the MTLDM, which achieved the lowest CRPS values across all of the pooling scales and prediction ranges. This result highlights the effectiveness of the MTLDM in learning-rich temporal dependencies and generating accurate and well-calibrated probabilistic predictions. The other algorithms (especially the TRDMv2) managed to narrow their performance gaps with that of the MTLDM as the pooling scale increased. Moreover, we observed a general trend of increasing CRPS value over time for all of the algorithms, also demonstrating that prediction becomes increasingly difficult as time progresses.
 \textcolor{black}{
 The CRPS indicators of the SW dataset at various scales are shown in Tables \ref{table6}. For a pooling scale of 1, the MTLDM's CRPS values are significantly lower than those of the other five algorithms at most of the time steps. Specifically, for the 80-minute prediction (16 time steps), the MTLDM's CRPS value is 0.1440, about 10\% lower than that of DiffCast (0.1600) and about 14\% lower than that of DGMR (0.1679). For the 5-minute prediction, the MTLDM's advantage is even more pronounced, with its CRPS value (0.0842) about 5\% lower than that of DiffCast (0.0887), about 13\% lower than that of DGMR (0.0973) and about 21\% lower than that of PySTEPS (0.1062).
 For pooling scales of 4 and 16, the MTLDM's CRPS values are significantly lower than those of the other algorithms at most time steps, particularly in the early stages. However, it's worth noting that TRDMv1 outperforms MTLDM at later time steps (from step 9-15 at scale 1, and from steps 10-15 at scales 4 and 16). The results demonstrate that the MTLDM achieves the lowest CRPS values at the majority of time steps, with its advantage being particularly evident in the early stages of the prediction. At the beginning of the prediction, the MTLDM's CRPS values are approximately 5-20\% lower than those of the best competing algorithms.
 Overall, the MTLDM demonstrates a clear advantage for this dataset at early time steps, while TRDMv1 shows competitive performance at later time steps. DiffCast generally performs better than DGMR, TRDMv2, and PySTEPS across different scales, especially at early time steps. The performances of all algorithms decrease as the prediction time increases, but the rate of decrease slows down in the later stages.}

 \textcolor{black}{The CRPS indicators of the MRMS dataset at various scales are shown in Tables \ref{table7}. For a pooling scale of 1 km, the performance comparison shows interesting patterns among the models. DiffCast achieves the best performance at the first time step with a CRPS of 0.0725, while MTLDM(oa) demonstrates superior performance across all subsequent time steps (2-16). As the pooling scale increases to 4 km and 16 km, all algorithms exhibit overall lower CRPS values, indicating improved prediction quality at coarser spatial resolutions. However, MTLDM(oa) consistently maintains the lowest CRPS values across all time steps at these larger scales, demonstrating its robust performance across different spatial resolutions. Among traditional methods, PySTEPS generally shows higher CRPS values than the deep learning approaches. DGMR performs better than TRDMv1 and TRDMv2 at early time steps, but this advantage diminishes in later prediction periods. DiffCast only outperforms MTLDM(oa) at the first time step at the 1 km scale, but falls behind at all other time steps and scales, showing a notable performance gap compared to MTLDM(oa).}

 Based on the analysis above, it is evident that our proposed method, MTLDM, which utilizes a divide-and-conquer strategy for precipitation prediction, significantly outperforms the latest existing methods in terms of both the CSI and CRPS indices.
 \begin{figure*}[!t]
	\centering
	\includegraphics[width=0.95\textwidth]{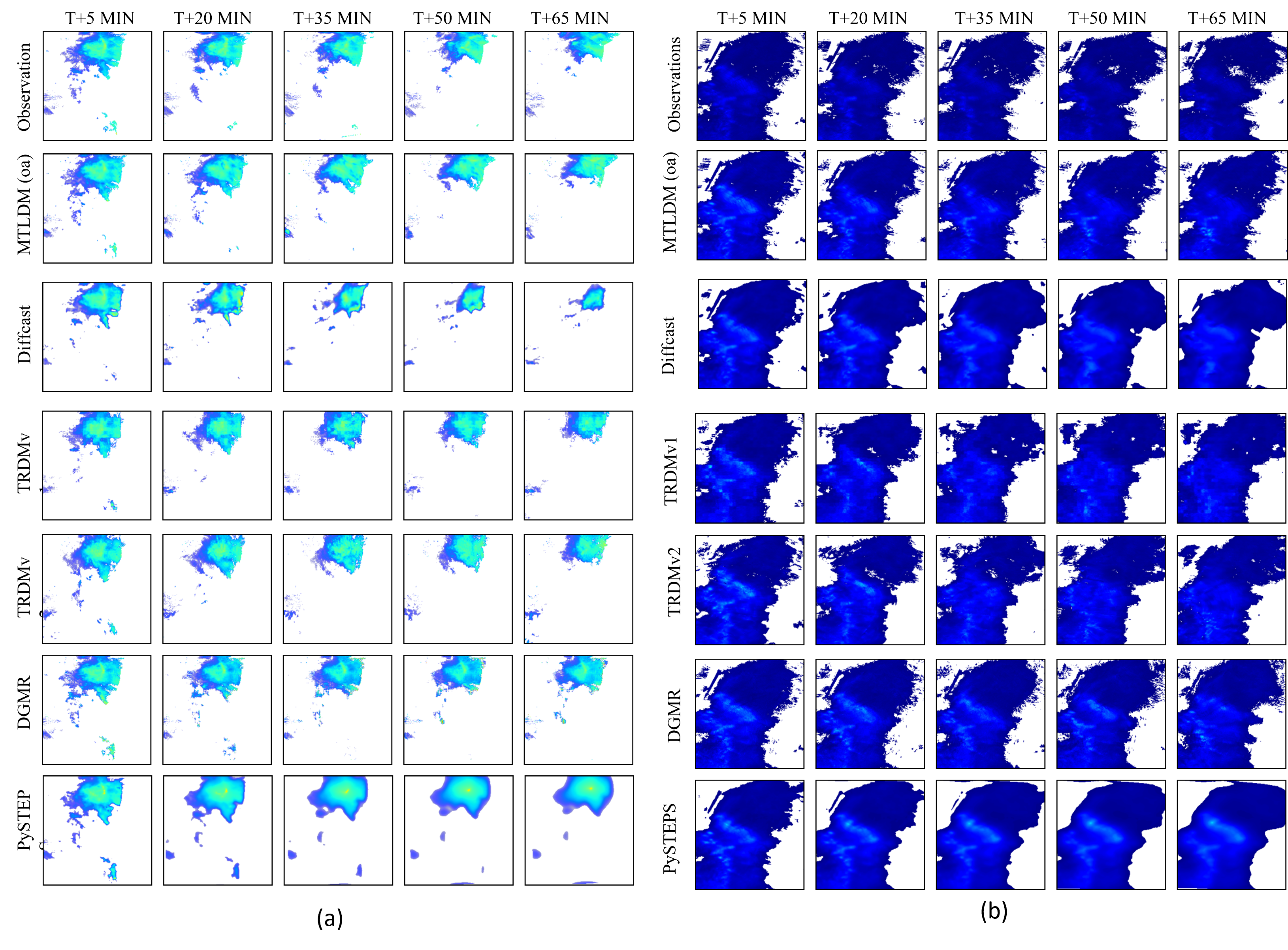}
	\caption{Prediction performance and visualization evaluation for the overall precipitation events. (a) The precipitation event that occurred on October 28, 2021, at 13:25:00 in the SW dataset. (b)The precipitation event that occurred on February 28, 2022, at 11:42:00 in the MRMS dataset. }
	\label{fig10}
\end{figure*}
\subsection{Evaluation based on Precipitation Events}

\begin{figure*}[htbp]
	\centering
	\includegraphics[width=0.95\textwidth]{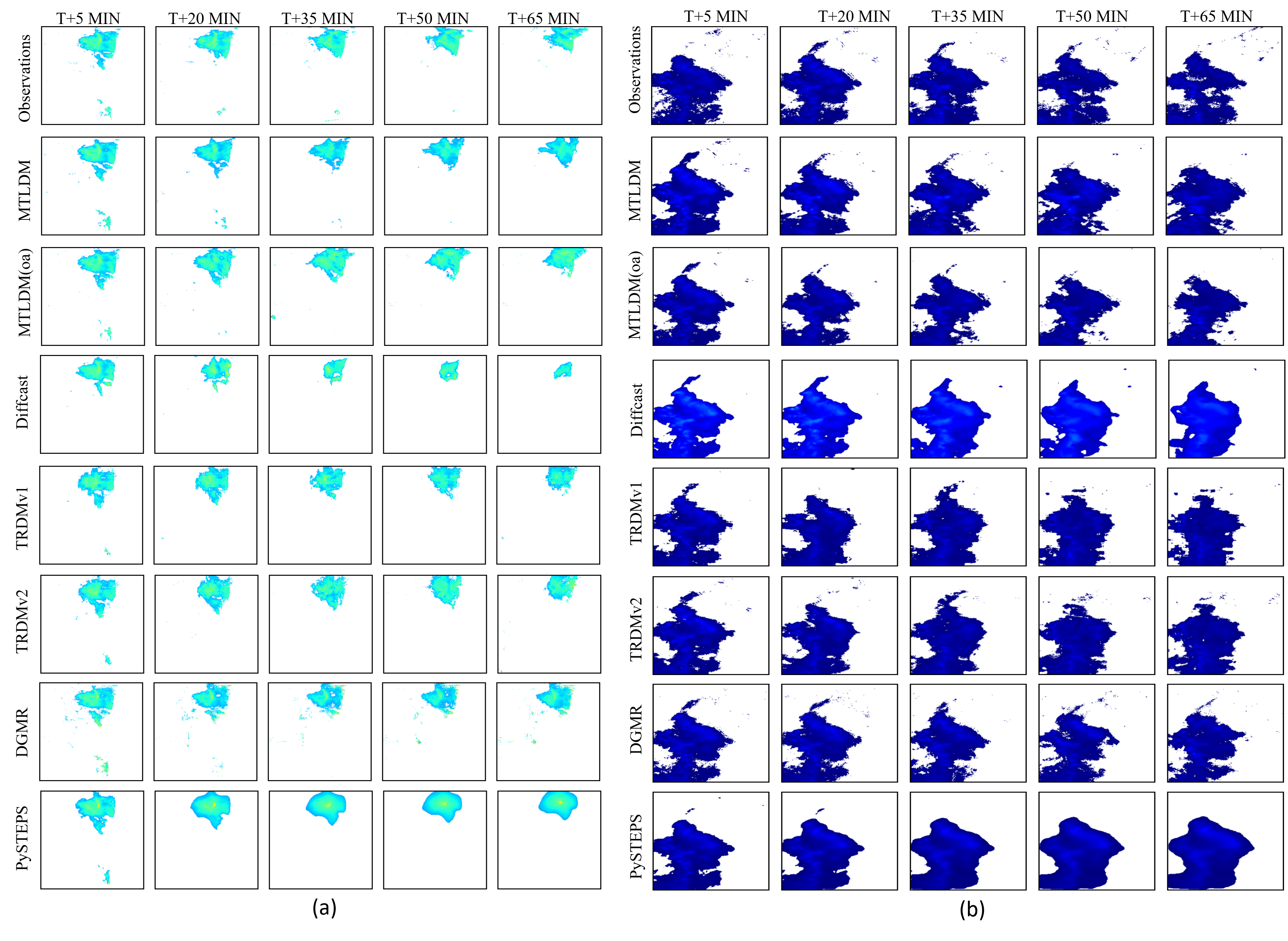}
	\caption{Prediction performance and visualization evaluation for the $\geq$1 mm/h precipitation event. (a)The precipitation event that occurred on October 15, 2021, at 13:25:00 in the SW dataset. (b) The precipitation event that occurred on February 28, 2022, at 11:42:00 in the MRMS dataset.}
	\label{fig13}
\end{figure*}
\begin{figure*}[htbp]
	\centering
	\includegraphics[width=0.95\textwidth]{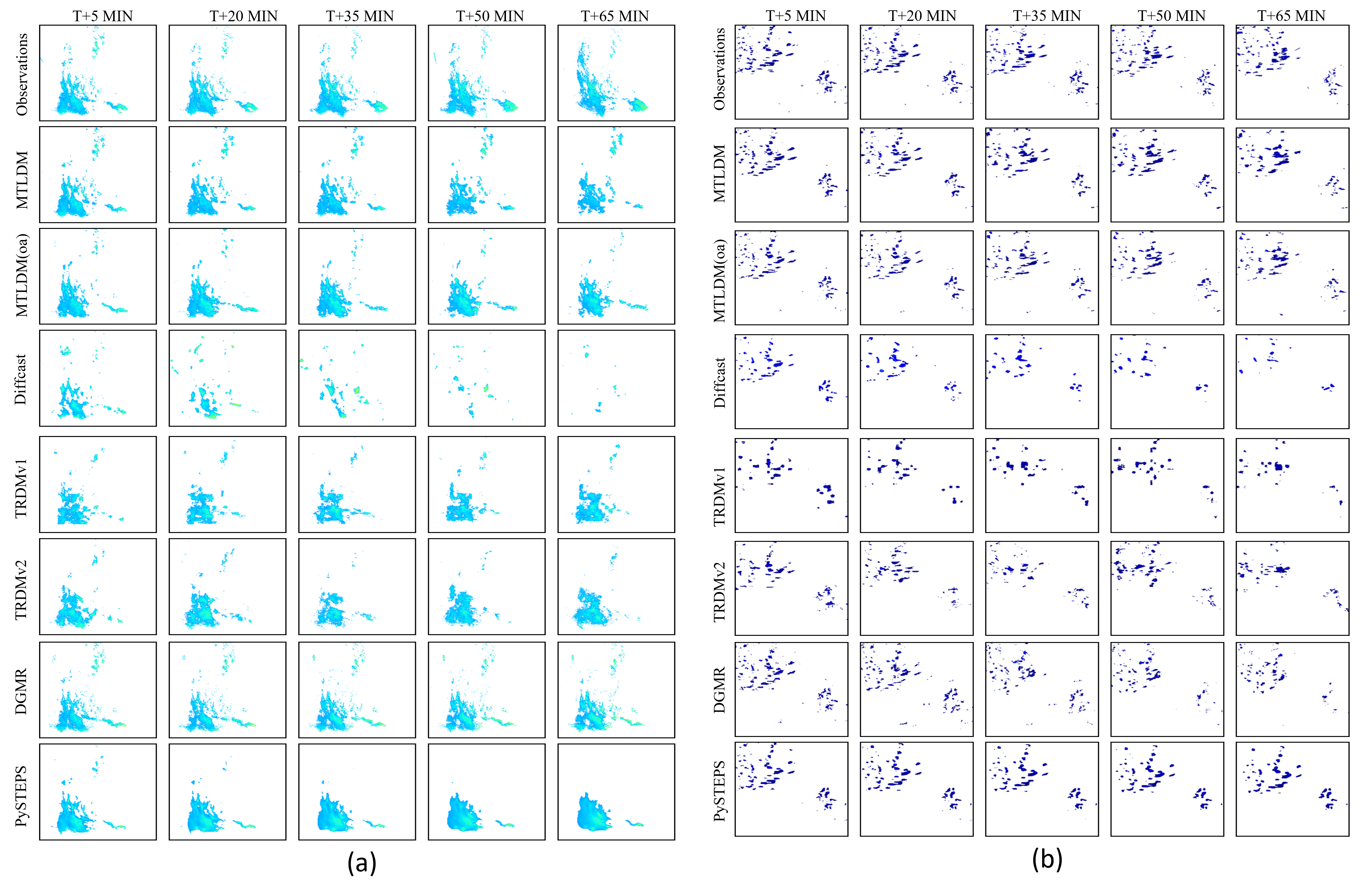}
	\caption{Prediction performance and visualization evaluation for the $\geq$1 mm/h precipitation event. (a)The precipitation event that occurred on November 02, 2021, at 02:08:00 in the SW dataset. (b) The precipitation event  that occurred on February 01, 2022, at 09:28:00 in the MRMS dataset.}
	\label{fig14}
\end{figure*}

In each dataset, we selected four precipitation events to evaluate the performance of the proposed prediction model. Two precipitation events were used to evaluate the prediction performance of the model for the overall precipitation (radar images are not segmented, see Figs \ \ref{fig-1}-\ref{fig10} ), and the remaining two precipitation events were used to evaluate the prediction performance of segmented radar images at different precipitation intensities ($\geq$ 1 mm/h).  Since precipitation of $\geq$ 4 mm/h is rare, we only make visual comparisons of the overall precipitation image and the precipitation forecast with a threshold of $\geq$ 1 mm/h (see Figs \ref{fig13} and \ref{fig14}).  During the evaluation process, the experimental results illustrate a comprehensive comparison of various precipitation forecasting models, benchmarked against the ground truth observations across different time steps, from T+5 minutes to T+65 minutes.  


 Overall, MTLDM demonstrates consistently superior performance in terms of both spatial accuracy and intensity prediction throughout the forecast window. Across all time steps, the spatial patterns and intensities predicted by MTLDM closely align with the observed precipitation, particularly in regions of higher intensity. This consistency over extended periods, ranging from T+5 to T+65 minutes, underscores MTLDM's robust predictive capabilities. In contrast, other models such as TRDMv1, TRDMv2, DGMR, and PySTEPS tend to display significant discrepancies, particularly at later time steps, where their predictions often diverge from the ground truth.

When examining the performance of the other models, TRDMv1/v2 and DGMR perform reasonably well at earlier time intervals but begin to falter as the prediction horizon extends beyond T+35 minutes. These models tend to over-smooth precipitation patterns or misalign spatially, especially in high-intensity regions, which results in less accurate predictions compared to MTLDM. PySTEPS shows even more pronounced difficulties, particularly in maintaining the spatial precision of precipitation forecasts over longer time frames. By T+50 and T+65 minutes, PySTEPS' predictions exhibit significant spreading of precipitation areas, resulting in a notable loss of intensity and spatial accuracy.

A critical strength of MTLDM lies in its ability to consistently capture the structure and intensity of precipitation over both short and extended periods. This is especially evident in higher-intensity regions, where MTLDM outperforms competing models by maintaining close proximity to the ground truth, even at longer time horizons. The ability to maintain accuracy in these high-intensity regions is particularly valuable for real-world applications, where such precision is essential for decision-making.

Further analysis focusing on precipitation events with intensities of $\geq$ 1 mm/h reinforces these findings. MTLDM continues to exhibit remarkable accuracy in both spatial and temporal domains. Across both the subfigures (a) and (b) of Figs \ref{fig10} and \ref{fig13}, MTLDM remains closely aligned with the ground truth, capturing the spread and intensity of precipitation with minimal deviation.MTLDM(oa) also has good prediction visual effects, but is slightly inferior to MTLDM in terms of the details of the predicted radar image. In contrast, other models, such as SSLDM, TRDMv1, TRDMv2, and DGMR, struggle to maintain this level of accuracy as the forecast window extends. While SSLDM and TRDMv1/v2 deliver relatively strong performance at shorter time intervals, they progressively lose accuracy at later time steps, with the predicted precipitation patterns becoming overly diffused or spatially shifted. \textcolor{black}{Due to limited network capacity, the precipitation results predicted by Diffcast show a trend of gradually weakening over time, resulting in significant differences between the long-term predicted precipitation areas and the actual observed values. This model limitation is clearly reflected in Figures \ref{fig10}.a and \ref{fig13}.a, where the deviation between the predicted values and the actual precipitation distribution increases significantly with the extension of the prediction time.}

The performance gap becomes more apparent in models such as DGMR and PySTEPS, where both models exhibit considerable difficulty in capturing the spatial intricacies of precipitation events with intensities exceeding 1 mm/h. DGMR's predictions, in particular, become increasingly inaccurate beyond T+35 minutes, with substantial differences in both spatial extent and intensity when compared to the ground truth. Similarly, PySTEPS exhibits substantial spreading of precipitation areas at later time steps, leading to significant discrepancies in both spatial and intensity accuracy.

In summary, MTLDM outperforms all other models in forecasting precipitation events, particularly those with higher intensities. Its ability to consistently maintain accuracy in both the spatial distribution and intensity of precipitation across multiple time steps, especially at extended time horizons, highlights its robustness and reliability for precipitation forecasting. Other models, while performing adequately at shorter time intervals, lack the precision and consistency demonstrated by MTLDM, particularly for high-intensity rainfall events. The superiority of MTLDM  across different forecast horizons makes it the most effective model for both short-term and extended precipitation predictions.

\section{Conclusions}

Accurately predicting extreme precipitation is critical for a variety of weather-dependent decision-making processes, but it remains a major challenge. Our multi-task latent diffusion model (MTLDM) provides an innovative solution to this problem by addressing the limitations of current deep learning models in capturing spatial details in radar images, especially in areas of high-intensity precipitation. MTLDM decomposes radar images into sub-images representing different precipitation intensities and predicts each component separately, thereby achieving better performance than existing models on multiple evaluation metrics and providing spatiotemporally consistent forecasts over forecast periods where existing methods often fail.
\textcolor{black}{
An important feature of MTLDM is the ability to decode the latent space. The main advantage of this approach is that it can directly operate on the latent space using a trained rainfall model and generate forecasts without additional training. However, this strategy of relying on the latent space also brings limitations: model performance is largely limited by the quality and accuracy of the latent space representation. If the latent space encoding is not sufficient to capture complex meteorological patterns, especially nonlinear changes in extreme weather events, it will limit the overall forecasting ability.
While MTLDM performs well in terms of accuracy and consistency in short-term forecasts, it remains challenging to predict heavy precipitation over longer forecast periods. Our evaluation shows that although MTLDM is a significant improvement over existing methods, the difficulty of accurately predicting extreme precipitation remains in all methods, and the model effect gradually decreases as the forecast time horizon increases.}
In addition, our study also highlights the limitations of conventional validation indicators, which may not fully reflect the practical application value of advanced models such as MTLDM. This suggests that new quantitative evaluation methods need to be developed to better adapt them to practical applications and ensure that the improvement in prediction performance can be translated into practical benefits for decision makers. We hope that this study will promote further integration of machine learning and environmental science, promote the development of new data, methods and evaluation techniques, and improve the prediction accuracy and practical utility of extreme weather event nowcasting.

\bibliographystyle{IEEEtran}
\bibliography{mybibfile} 

\end{document}